\documentclass[preprint,11pt, authoryear]{elsarticle}
\usepackage{lab}
\usepackage{soul}
\usepackage{makecell} 
\usepackage{algorithm}
\usepackage[colorlinks]{hyperref}
\usepackage{adjustbox}
\biboptions{authoryear}
\usepackage{listings}
\usepackage{wrapfig} 
\usepackage{amssymb}

\journal{International Journal of Forecasting}
\begin{document}

\begin{frontmatter}

\title{Interpretable Water Level Forecaster with Spatiotemporal Causal Attention Mechanisms}
\author[1]{Sungchul Hong}
\ead{shong@changwon.ac.kr}
\author[2]{Yunjin Choi}
\ead{ycstat@uos.ac.kr}
\author[2]{Jong-June Jeon\corref{cor}}
\ead{jj.jeon@uos.ac.kr}

\cortext[cor]{Corresponding author}
\affiliation[1]{organization={Department of Statistics, Changwon National University},
            addressline={20 Changwondaero, Uichang-gu}, 
            city={Changwon},
            postcode={51140}, 
            country={South Korea}}

\affiliation[2]{organization={Department of Statistics, Unversity of Seoul},
            addressline={163 Seoulsiripdaero, Dongdaemun-gu}, 
            city={Seoul},
            postcode={02504}, 
            country={South Korea}}

\begin{abstract}
Accurate forecasting of river water levels is vital for effectively managing traffic flow and mitigating the risks associated with natural disasters. 
This task presents challenges due to the intricate factors influencing the flow of a river.
Recent advances in machine learning have introduced numerous effective forecasting methods. 
However, these methods lack interpretability due to their complex structure, resulting in limited reliability.
Addressing this issue, this study proposes a deep learning model that quantifies interpretability, with an emphasis on water level forecasting. 
This model focuses on generating quantitative interpretability measurements, which align with the common knowledge embedded in the input data.
This is facilitated by the utilization of a transformer architecture that is purposefully designed with masking, incorporating a multi-layer network that captures spatiotemporal causation.
We perform a comparative analysis on the Han River dataset obtained from Seoul, South Korea,  from 2016 to 2021. 
The results illustrate that our approach offers enhanced interpretability consistent with common knowledge, outperforming competing methods and also enhances robustness against distribution shift.
\end{abstract}

\begin{highlights}

    \item We introduce a novel deep learning architecture that features enhanced interpretability, with a specific emphasis on its application with real-world datasets in water level forecasting.
    
    \item Our proposed model is capable of guiding the outputs to align with the input common knowledge. This input, comprising common knowledge, is encoded through a multi-layer network framework that captures the underlying spatiotemporal structure.
 
    \item In the real-world water level data, our proposed model outperforms the state-of-the-art in terms of interpretability while enhancing robustness to the distribution shift.
\end{highlights}


\end{frontmatter}

\section{Introduction}\label{sec:intro}

Deep learning prediction models are widely employed across diverse industries and sectors, encompassing domains such as finance, healthcare, and logistics management \citep{chatigny2021spatiotemporal, sezer2020financial, Avati2017ImprovingPC, Kaneko2016ADL}. 
However, evaluating the reliability of neural network predictions presents a substantial challenge, largely attributed to the intricate nature of interpreting the results. This challenge arises from the complicated structures of neural networks that incorporate multiple compositions of nonlinear functions.
As a result, diagnosing and addressing issues pertaining to output reliability during the training phase becomes challenging. The lack of interpretability and direct accountability for model outputs undermines their trustworthiness, potentially discouraging the further utilization of neural network models.

In the same light, the importance of model interpretability becomes vital in the context of developing a river's water level forecasting model.
The behavior of river water levels is affected by the laws of physics across both space and time, giving rise to distinct guiding principles.
For instance, the upstream water flow within the defined spatial framework exerts a notable influence on the downstream flow at any given moment. Furthermore, the dynamics of water flow, as elucidated by the principles of fluid dynamics, play a crucial role in accounting for temporal features.
Previous studies have emphasized the importance of the forecasting model adhering to established physical laws or features, particularly with regard to spatial and temporal dependencies \citep{wu2020complexity, Fang2020PredictingFS}.
Even in situations where the neural network model outperforms human experts in forecasting the given dataset, there is still a possibility that the model's mechanism may exhibit general unreliability.
Hence, when the results generated by the forecasting model diverge from the perspectives of domain experts or established principles of physics, it becomes imperative to conduct further analysis. In such cases, the interpretability of the results proves to be highly advantageous.

In this study, our focus lies in addressing the challenge of constructing a deep learning architecture capable of embodying spatial and temporal dependencies simultaneously while also delivering interpretable results. Specifically, our emphasis is directed toward forecasting the water levels of a river.
The model we propose produces probabilistic forecasts for the water level at a specific river location, providing a quantified measure of output uncertainty.
In our proposed approach, we introduce a novel attention framework that accommodates a multi-layered network structure, effectively encapsulating the inherent spatiotemporal nature of a provided dataset.
Our model introduces two attention weights designed to capture the correlations between spatial and temporal features. 
The construction of each attention weight involves strategic maskings to align with our prior knowledge, encompassing factors such as temporal causality and the laws of spatial physics.
This approach enables the training of a model that considers a causal structure and generates forecasts within a constrained model space based on this underlying causal structure.
Our proposed architecture does not necessitate a complete identification of the causal structure, and partial knowledge can still be encapsulated.
The employment of this approach improves forecasting performance.
Moreover, our proposed method attains a greater level of flexibility compared to the existing spatiotemporal forecasting method.
This advanced flexibility arises from the capability of our method to utilize spatially heterogeneous predictors.
This ability is facilitated by summarizing spatial features through feature-specific embedding layers.

\subsection{Related Work}

Our proposed approach aims to utilize deep learning for probabilistic forecasting of the water level through spatiotemporal modeling, emphasizing interpretability. We review relevant literature, focusing on the three key aspects of probabilistic forecasting, interpretable AI, and spatiotemporal modeling.

\textbf{Probabilistic forecasting.}  As opposed to point estimates, a probabilistic forecaster generates more informative results regarding target variables. 
This includes providing conditional distributions or multiple quantiles that prove invaluable for decision-making. 
Such forecasting is considered challenging due to its complexity. 
Yet, despite its difficulty, it holds considerable utility in various fields, including water level forecasting. Particularly, its capacity to quantify the risk linked to rare events like floods carries significant importance.
In this paper, we introduce a novel probabilistic forecasting method founded upon deep learning architecture. The proposed method yields forecasts for multiple quantiles. 

Recently, there have been notable advancements in the realm of probabilistic forecasting, particularly within the context of deep learning-based methods. State-of-art models leverage a variety of features, encompassing historical, categorical, and even prospective information, like dates and weekly projected weather, to achieve accurate forecasts of the target variable.
Various methods, such as DeepAR \citep{salinas2020deepar}, MQ-RNN \citep{wen2017multi}, and Temporal Fusion Transfomer (TFT, \cite{lim2021temporal}), have been introduced, gaining widespread adoption across diverse domains due to their powerful performance capabilities. 
Among these, DeepAR employs a seq2seq \citep{cho2014learning} architecture to estimate parameters of the target distribution at future time points.
Although its implementation is relatively straightforward, DeepAR exhibits certain limitations, primarily that the target distribution is expected to conform to an assumed parametric family of distributions.
MQ-RNN and TFT do not rely on a specific distribution assumption and align with our proposed approach, given their shared objective of forecasting multiple quantiles for the target distribution.
MQ-RNN utilizes feedforward networks and gains computational efficiency and learning stability in the process.
TFT, rooted in transformer architecture \citep{Vaswani2017AttentionIA}, adeptly handles complex types of input variables, including static variables and variables known for future time points. This attribute leads to enhanced performance, enabling its effective application across diverse domains \citep{wu2022interpretable, zhang2022temporal}.
Our proposed method extends the framework of TFT, expanding its capabilities further.

\textbf{Interpretable AI.} 
Owing to the increasing demand for comprehending the outcomes of complicated models to ensure their reliability, the pursuit of Interpretable AI has gained considerable popularity across diverse domains, including water level forecasting \citep{Ding2020InterpretableSA, castangia2023transformer}.
Many of these applications in the realm of Interpretable AI have been constructed based on the framework of TFT, showcasing enhanced interpretability alongside notable forecasting performance \citep{civitarese2021extreme, mu2023icetft}.

TFT, a model built upon the transformer architecture, achieves enhanced interpretability through the quantification of variable importance. 
Despite the TFT's capability to provide interpretability, the resulting interpretations might not fully align with the innate relationships between variables, such as temporal changes or spatial dependencies.
This limitation may arise from its fundamental design.
Specifically, TFT attains its interpretive strength by integrating a variable selection network and an attention mechanism, which yields quantified evaluations of variable importance.
In this setup, the variable selection networks are placed independently at each time point's input layer. 
As a result, variable importances for each time point are calculated separately, potentially risking the oversight of innate relationships among variables.
In our proposed method, we enhance TFT by integrating masking techniques that encode the presumed interconnections between variables.
This integration results in interpretation outcomes that are consistent with the innate interconnectedness among variables.

\textbf{Spatiotemporal modeling.} 
In water level forecasting, the careful construction of a spatiotemporal stochastic framework for water flow is imperative. This entails the simultaneous consideration of both temporal dynamics and spatial modeling, thereby contributing collectively to the effective management of hydrological time series data.
However, capturing these simultaneous effects presents a challenge, with several existing methods addressing it sequentially, such as through a two-step approach. In the two-step approach, the initial focus lies on filtering temporal dependencies, followed by the construction of spatial dependencies.
For illustration, during the initial stage, the temporal filter--comprising components such as autoregressive models, wavelet transformation, and empirical mode decomposition--captures temporal features \citep{Yadav2017AHW, Wu2021GroundwaterLM}. Subsequently, the filtered temporal features from multiple sites are aggregated across a spatial domain utilizing nonlinear models, including the support vector machine, neural network model, and neuro-fuzzy system \citep{Ruslan2014FloodWL, Yadav2017AHW}.
This approach may have a limitation in effectively capturing the interplay between spatial and temporal dependencies due to the absence of concurrent consideration for their intertwined effects.

Recently, deep learning-based models have been spotlighted in the field of hydrological time series forecasting, owing to their capability to integrate simultaneous spatiotemporal modeling in a straightforward manner.
Mainly, there are two approaches for constructing spatiotemporal deep learning models, outlined as follows. The first approach involves constructing a model structure in a constrained manner, thereby customizing the model architecture to a specific spatiotemporal structure of a given dataset \citep{Ding2020InterpretableSA, liu2022directed}.
This approach has a limited scope and is applicable only to specific datasets due to its tailored architecture.
The second approach involves the utilization of a graph neural network (GNN), which is a neural network capable of handling graph-structured data.
In their study, \cite{Deng2022ASG} presented a GNN-based method focusing on river network analysis. This method involves capturing spatial dependencies through a graph convolution network, as well as extracting temporal patterns through the application of either a recurrent neural network (RNN), temporal attention mechanism, or temporal convolution network.
In addition to hydrologic time series modeling, GNN-based methods find applications in diverse domains where spatiotemporal dynamics are inherent. For instance, these methods are utilized in predicting pedestrian trajectories \citep{Zhou2021ASTGNNAA} and forecasting traffic patterns \citep{Roy2021SSTGNNSS}.
However, these methods present an increased challenge in terms of delivering interpretability, as they integrate complicated models such as GNNs and RNNs. 
To the best of our knowledge, interpretable forecasters based on the GNN remain underdeveloped. 
Additionally, these methods lack flexibility in terms of accommodating diverse covariate forms and do not support heterogeneous types of covariates across different sites on the spatial domain.

In contrast to the previously discussed approaches, we introduce a general method that simultaneously takes into account spatiotemporal dependencies and accommodates diverse covariates across sites.
In modeling spatiotemporal structures, our proposed method utilizes a simple architecture relative to GNNs. This maintains enhanced interpretability as compared to TFT, ensuring that the interpretation results align with conventional knowledge.

The remainder of this paper is organized as follows. Section 2 introduces the dataset of interest in this paper and model assumptions dominated by physics law. Section 3 explains the proposed model, focusing on novel attention mechanisms. Section 4 shows the numerical result from real data analysis, which provides explainable quantities for understanding the dataset. Concluding remarks and limitations of this study follow in Section 5. 

\section{Preliminary}
\subsection{Dataset} \label{sec:data}

In this study, our main focus is on forecasting the water level of Jamsu Bridge, an important structure located in Seoul, South Korea. Spanning across the Han River, the Jamsu Bridge serves as a crucial link between the bustling business districts on the north and south sides. One of the distinctive features of the Jamsu Bridge is its intentionally low elevation, which was designed to be at 2.7 meters during its construction in 1976. 
This unique attribute sets it apart from other bridges in the vicinity, as they are typically located $16$ to $20$ meters above the water level.
The Jamsu Bridge's unique low elevation also makes it highly susceptible to flooding \citep{lee2017operational}. In 2020, the bridge was completely submerged for 232 consecutive hours in 2020.\footnote{\url{https://www.codil.or.kr/viewDtlConRpt.do?gubun=rpt&pMetaCode=OTKCEC210998}}
Despite this vulnerability, the bridge remains under high demand, handling a substantial flow of 22,673 cars per day in 2020.\footnote{\url{https://news.seoul.go.kr/traffic/files/2012/02/6058855d14fa49.45283783.pdf}} 
As a result, the water level of the Jamsu bridge draws considerable and distinct attention during every flood season. Its accurate forecasting has become crucial to ensure safety and maintain a smooth traffic flow on the bridge. 
This necessity strongly motivates our study to concentrate specifically on this bridge.

\begin{figure*}[t]
    \centering
    \includegraphics[width=0.92\linewidth]{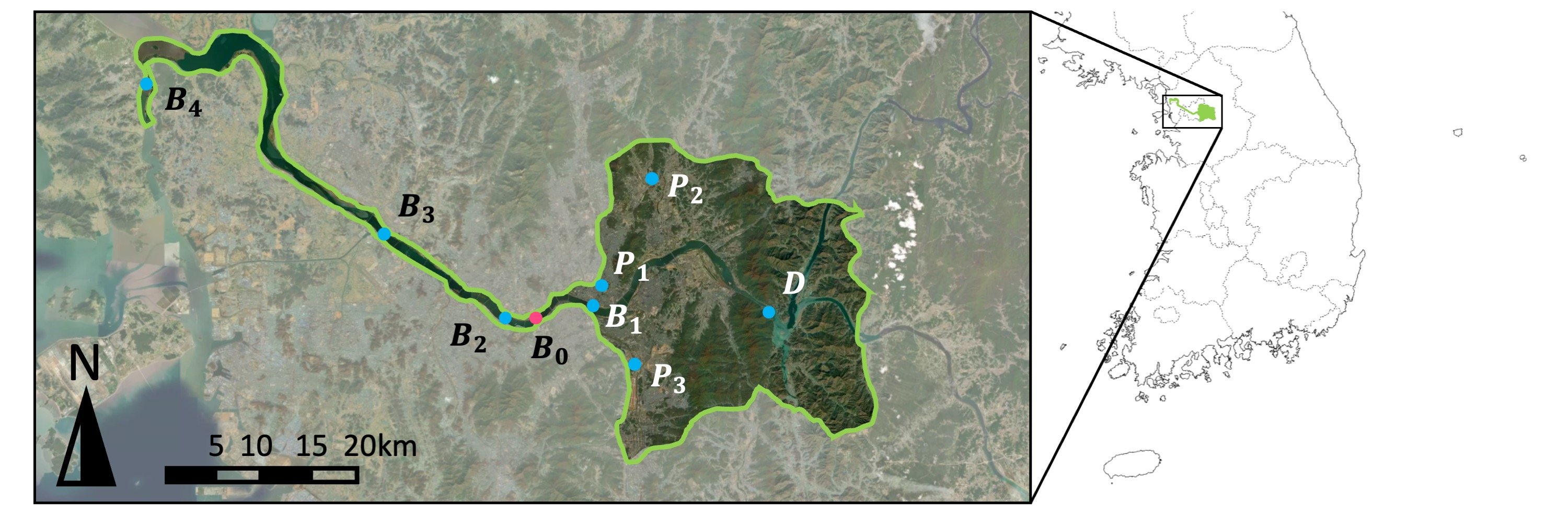}
    \caption{The satellite image of Han River, Seoul, South Korea with highlights on the area of interest. The blue points indicate the observatories at which observations are collected, and the red point denotes our target site, the Jamsu bridge $B_0$. The river flows from east to west, emptying into the sea.}
    \label{fig:han_river}
\end{figure*}

The data used for forecasting the water level of the Jamsu Bridge is collected from six observatories located along the Han River, as well as three additional meteorological observatories. The dataset spans from 2016 to 2021, inclusive. The observatories on the river are Paldang Dam ($D$), Cheongdam Bridge ($B_1$), Hangang Bridge ($B_2$), Haengju Bridge ($B_3$), Ganghwa Bridge ($B_4$), and Jamsu Bridge ($B_0$). 
The observatories involved in the study collect a diverse range of covariates, with specific types depending on the type of each observatory.
Specifically, the Paldang Dam observatory ($D$) collects time series data comprising water level (WL), inflow (IF), outflow (OF), storage (STR), and joint usage storage (JUS) measurements.
The observatories located on bridges ($B_1, B_2, B_3, B_4$) collect water levels (WL) and flow (FL). 
In addition, the meteorological observatories ($P_1$, $P_2$, $P_3$), located near the Han River, collect precipitation data. At the target site Jamsu Bridge ($B_0$), only water level data is collected.
Covariates collected at specific sites are denoted by pairing the covariate name with the corresponding site name in parentheses. For example, the water level measured at bridge $B_1$ is denoted as WL $(B_1)$. In the case of precipitation covariate, we use the site name alone, as meteorological observatories are exclusively associated with the precipitation variable.
Moreover, in our analysis, we incorporate temporal variables such as month, day, and hour.
All variables, excluding the temporal variables, are provided on an hourly basis.
Figure \ref{fig:han_river} presents a map of the Han River area, indicating the locations of the observatories.
The summary statistics for the variables used in the analysis are available in the Appendix.

The Han River is the second longest river in South Korea, traversing the city of Seoul before reaching the West Sea. Spanning a length of 508 kilometers and encompassing a basin area of 35,770 square kilometers, the Han River holds considerable hydrological and geographical importance. The average discharge at the Hangang Bridge, which is a prominent landmark along the river, is about 613 cubic meters per second. 
Due to its vast size and its intricate relationship with numerous factors, developing a water level forecasting model that encompasses all these factors can present a substantial challenge. 
For instance, in the Han River, the water level in the upper stream can be subject to the influence of downstream conditions, contradicting our intuitive understanding and basic principles of physics. This phenomenon can be attributed to the proximity of the Han River to the sea, which exposes it to tidal effects. As a result, tides cause an increase in the water level downstream, subsequently elevating the water level in the upper stream as well \citep{park2017reconsideration}.
An example of such a case is the Ganghwa Bridge, located downstream of the Jamsu Bridge. Hence, the water level of Ganghwa Bridge serves as a crucial predictor for forecasting the water level of Jamsu Bridge \citep{Jung2018Prediction}.
In this context, the integration of domain expert knowledge encompassing fundamental principles of physics and empirical findings becomes essential. 
Our proposed method is specifically designed to incorporate such essential domain expert knowledge during its construction.

\subsection{Modelling Spatiotemporal Causality via Multilayer Network} \label{sec:mscmn}
In capturing the spatiotemporal structure of the dataset, we utilize the multilayer network framework.
The multilayer network is a useful tool for modeling a pattern across variables with a hierarchical structure \citep{kivela2014multilayer}, such as biomedicine \citep{hammoud2020multilayer} and community detection \citep{Huang2020ASO}. In the context of spatiotemporal structure, 
\cite{choi2022capturing} employed a multilayer network approach to capture the patterns of the bike-sharing system. This method allows for the simultaneous consideration of both time and space factors.
As in the previous studies, our approach employs a multilayer network to facilitate the learning of spatiotemporal variables.
In our framework, spatial causality is modeled as a directed graph on each layer, where each layer corresponds to a particular hour of the day. 
Additionally, temporal causality is captured by directed edges that connect the layers.

In the construction of the multilayer network structure, each layer consists of four nodes, all having the same network structure.
Each node is associated with a predetermined group of observatories, based on their specific characteristics. These groupings are as follows: the meteorological observatories group $C_1= \{P_1, P_1, P_3\}$, the first group of bridges $C_2=\{B_4\}$, the dam group $C_3= \{D\}$, and the second bridge group $C_4=\{B_0, B_1, B_2, B_3\}$. Within these groupings, the bridges are separated into two distinct groups of 
$C_2 = \{B_4\}$ and
$C_4 = \{B_0, B_1, B_2, B_3\}$. 
This separation is due to the unique role of $B_4$, which is the Ganghwa bridge. Being located downstream and influenced by the tide, $B_4$ impacts the water level upstream in turn, including the other bridge group $C_4= \{B_0, B_1, B_2, B_3\}$ \citep{Shin2005TheSpatial, park2017reconsideration}. 

In our multilayer network, the nodes on the $t$-th layer are denoted by $v_{s,t}$ where $s$ corresponds to the predetermined clusters $C_s$ with $s \in \mathcal{S} = \{1,2,3,4\}$. The set of nodes on the $t$-th layer are denoted by $\mathcal{V}_t = \{v_{1,t}, v_{2,t}, v_{3,t}, v_{4,t}\}$.
Each $t \in \mathcal{T} =\{1, ..., T\}$ corresponds to a specific hour within a duration of $T$ consecutive hours.
The edges connect these nodes in a directed manner, traversing both inter and intra layers, and encoding spatiotemporal causal structures. 
The construction of these edges follows Assumptions \ref{assum} and \ref{assum2}, which will be introduced in the subsequent paragraphs.
Our proposed multilayer network, denoted by $\mathcal{G}$, is defined as a tuple of three sets, which are a set of nodes $\mathcal{V}$, a set of edges $\mathcal{E}$, and a set of layers $\mathcal{T}$:
\[
\mathcal{G} = (\mathcal{V}, \mathcal{E}, \mathcal{T}),
\]
where $\mathcal{V} = \underset{t \in \mathcal{T}}{\bigcup} \mathcal{V}_t$.
The edge structures of $\mathcal{E}$ are specified in Assumption \ref{assum} and Assumption \ref{assum2}, encoding spatiotemporal causality. 
In our graph, all edges are directed, representing causal relationships. An edge denoted as $(v_{s,t}, v_{s',t'}) \in \mathcal{E}$ indicates a causal relationship, where the occurrence of the former node $v_{s,t}$ is a  \textit{cause} of the occurrence of the later node $v_{s',t'}$. This relationship is also represented by $v_{s,t} \rightarrow v_{s',t'}$. 
Assumption \ref{assum} represents temporal causality, and Assumption \ref{assum2} represents spatial causality. 

\begin{assumption} \textup{(Temporal Causality)} \label{assum}
For $s,s' \in \mathcal{S}$ and  $t, t' \in \mathcal{T}$, the edges in the multilayer network $\mathcal{G} =(\mathcal{V}, \mathcal{E}, \mathcal{T})$ satisfies the following conditions:
\ben
    \item For $t \neq t'$, $(v_{s,t}, v_{s', t'}) \in \mathcal{E}$ holds only if $s = s'$
    \item Suppose that $t \leq t'$, then $(v_{s, t'}, v_{s, t}) \notin \mathcal{E}$.
    \item For $s \neq s'$, $(v_{s, t}, v_{s', t}) \in \mathcal{E}$ if and only if  $(v_{s, t'}, v_{s', t'}) \in \mathcal{E}$.
\een
\end{assumption}
Specifically, Assumption \ref{assum}.1 represents self-temporal causality, where the same node at different time points directly influences itself in a temporal manner.
Assumption \ref{assum}.2 states the irreversibility of time, indicating that only a preceding status can impact a later status, while the reverse does not hold.
Assumption \ref{assum}.3 implies homogeneity in spatial causality, indicating that the spatial causality structure remains consistent across time.

\begin{figure}[!t]
    \centering
    \includegraphics[width=\linewidth]{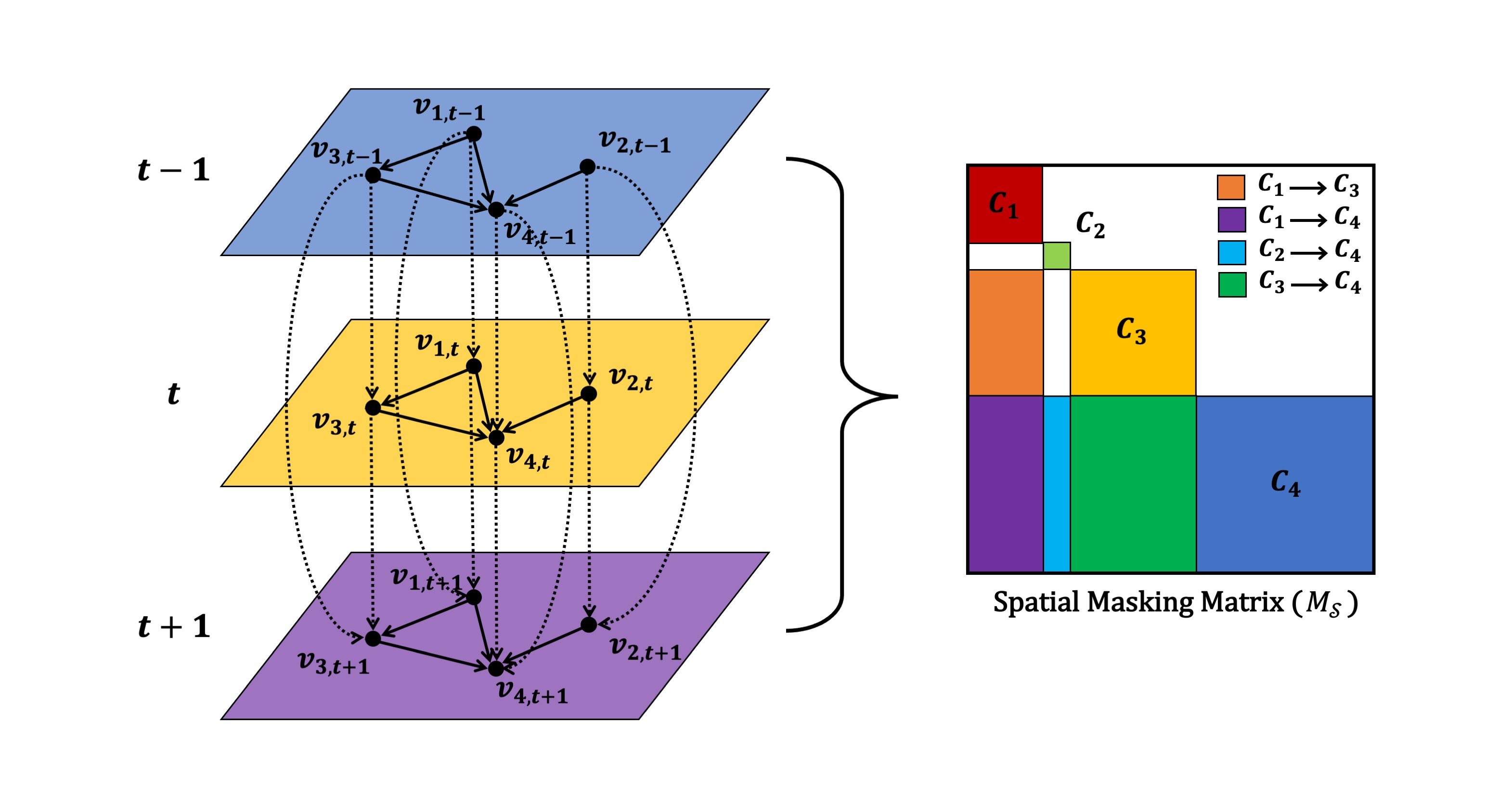}
    \caption{
    Left: The multilayer network $\mathcal{G}$, illustrating the spatiotemporal causal relations represented for three consecutive hours: $t-1$, $t$, and $t+1$. Each layer shares the same spatial causal relations, and every node is connected to a corresponding node in the subsequent layer, belonging to the same cluster.
   Right: Visualization of the spatial masking matrix $M_\mathcal{S}$, with colored blocks representing encodings of connected edges.}
    \label{fig:ml-graph}
\end{figure}

\begin{assumption}[Spatial Causality] \label{assum2}
 $(v_{s,1}, v_{s',1}) \in \mathcal{E}$, if $(s,s') \in \{(1,3), (1,4), (2,4), (4,3)\}$. Otherwise, $(v_{s,1}, v_{s',1}) \notin \mathcal{E}$.
\end{assumption}

Assumption \ref{assum2} embodies spatial causality based on the prior knowledge of domain experts. 
Omitting the time index, the assumed spatial causal relations can be represented as follows: 
\bea
v_{1} \rightarrow v_{3}, \ v_{1} \rightarrow v_{4}, \  v_{2} \rightarrow v_{4},  \ v_{3} \rightarrow v_{4}. \nonumber
\eea
Specifically, the node $v_1$ is associated with $C_1$, which represents precipitation measured at three meteorological observatories, $P_1$, $P_2$, and $P_3$ ($C_1 = \{P_1, P_2, P_3\}$). In our model, this node acts as a globally influential variable, influencing measurements at nearby observatories $v_3$ (Paldang Dam) and $v_4$ (the bridge cluster).
The node $v_2$ corresponds to the Ganghwa Bridge which is encoded as $C_2 = \{B_4\}$. As previously stated, despite its downstream location, the Ganghwa Bridge exerts an influence on the water level of the bridge cluster, denoted by $v_4$. This influence is attributed to the tidal characteristics of the river \citep{park2017reconsideration} and has been recognized as a crucial factor in previous studies forecasting the water level of the Jamsu Bridge \citep{Jung2018Prediction}.
The node $v_3$ is associated with $C_3 = \{D\}$, representing the Paldang Dam located in the upper stream of the river. The Paldang Dam directly influences the Han River's water level as a whole, including $v_4$ (the bridge cluster).
The node $v_4$ is associated with $C_4= \{B_0, B_1, B_2, B_3\}$, representing the bridge cluster, including our target bridge, the Jamsu Bridge ($B_0$). In our causal model, $v_4$ is influenced by all other nodes, including participation ($v_1$), the bridge closely connected with tidal patterns ($v_2$), and the upstream dam ($v_3$).
Table \ref{tab:causality} provides a summary of the spatial causality in our model. The overall multilayer structure of our network $\mathcal{G}$, displaying Assumptions \ref{assum} and \ref{assum2}, is exhibited in Figure \ref{fig:ml-graph}.

\begin{table}[t!]
    \centering
    \caption{The descriptions of the spatial causalities in Assumption \ref{assum2}.}
    \begin{adjustbox}{width=\textwidth}
    \begin{tabular}{c l} 
    \Xhline{1.5pt}  
    Causation &  Description  \\ \hline 
    $v_1 \rightarrow v_3$ & The precipitation affects the variables of the dam. \\
    $v_1 \rightarrow v_4$ & The precipitation affects the variables of the bridge cluster. \\
    $v_2 \rightarrow v_4$ & The variables of the Ganghwa bridge affect the water levels and flows of the bridge cluster. \\
    $v_3 \rightarrow v_4$ & The variables of the dam affect the water levels and flows of the bridge cluster. \\
    \Xhline{1.5pt}   
    \end{tabular}
    \end{adjustbox}
    \label{tab:causality}
\end{table}

\subsection{Attention Mechanism}

Attention is a mechanism that enables a neural network to selectively focus on informative parts of input features while making predictions. By assigning higher weights or importance to specific elements within the input sequence, it effectively captures dependencies and relationships within the input.
In our proposed method, we utilize the attention mechanism to ensure the model follows the predefined causality structure represented as a multilayer network structure in Figure \ref{fig:ml-graph}.

Specifically, the attention mechanism can be defined as a mapping from a sequence to another sequence. Given a $t\times d$ matrix $V$ representing a sequence consisting of $t$ ordered elements, each comprising $d$ dimensions, the attention mechanism outputs a $t \times d$ matrix $V'$. This output represents a sequence of length $t$, with each element having $d$ dimensions, in a similar manner. 
Along with the input value sequence $V$, the attention mechanism incorporates two additional matrices, $Q \in \mathbb{R}^{t \times d}$ and $K \in \mathbb{R}^{t \times d}$, as inputs. These additional matrices are associated with the query and key sequences at each layer within the neural network context.
The attention for $V$ is defined as
\bea \label{eq:attention}
\mbox{Attention}(Q,K,V) = \mbox{softmax}\left( Q K^\top /\sqrt{d} \right) V,
\eea
where $\mbox{softmax}$ is a row-wise softmax function, and $\text{softmax}(U)$ represents the matrix obtained by applying softmax along the rows of matrix $U$. The $i$-th row of $\text{softmax}(U)$ is computed as $\exp(U_{i\cdot})/\sum_j \exp(U_{ij})$, where $U_{i\cdot}$ represents the $i$-th row of matrix $U$.
In attention mapping \eqref{eq:attention},
\begin{equation}
\label{eq:att_weight}
    A = \mbox{softmax}\left( Q K^\top /\sqrt{d} \right)
\end{equation}
 is called the attention weight, which assigns weights to the elements in the input value sequence $V$. 
Denoting the element at the intersection of the $i$-th row and $j$-th column of matrix $A$ as $a_{ij}$, and the $i$-th row of matrix $V'$ as $V'_{i \cdot}$ (and similarly for other matrices), it is straightforward to verify that $V'_{i \cdot}$ is obtained as the weighted average of the rows of matrix $V$, with the weights being given by the elements of matrix $A$:
\[
V'_{i \cdot} = \sum_{j=1}^{t} a_{ij} V_{j \cdot }.
\]
These weights, represented by $a_{ij}$, indicate the importance of elements in $V$.

The attention weight matrix $A$ in \eqref{eq:att_weight} can be extended to incorporate the \textit{masking} technique by utilizing a predefined masking matrix $M$ as follows
\bea
A = \mbox{softmax}\left( Q K^\top /\sqrt{d}  \odot M \right), \label{eq:att_masking}
\eea
where $\odot$ is the elementwise product operator and 
the element of $M$ is either $1$ or $-\infty$.
In the context of the attention mechanism, masking refers to a technique, used to selectively hide or ignore certain elements or positions in the input data.
Through the application of masking, we gain the ability to identify or control particular characteristics of the trained features.
Specifically, when setting $M_{ij}$ to minus infinity, the corresponding attention weight $a_{ij}$ becomes zero. This effectively excludes $V_{j \cdot}$ from contributing to the construction of the feature $V'_{i \cdot }$ considering that $V'_{i \cdot } = \sum_{j=1}^t a_{ij} V_{j \cdot }$.
In our study, the masking technique facilitates the embodiment of the designed causal structure 
in Section \ref{sec:mscmn}, whcih is represented as a multilayer network $\mathcal{G} = (\mathcal{V}, \mathcal{E}, \mathcal{T})$. By setting the elements of $M$ to $-\infty$ that correspond to edges that are not in $\mathcal{E}$ and assigning the value $1$ to elements corresponding to the existing edges in $\mathcal{E}$, the causal structure $\mathcal{G}$ is appropriately represented. This approach ensures the preservation of the designed causal structure, as described in Section \ref{sec:mscmn}.

In our study, we specifically focus on self-attention, which is a form of attention mechanism. Unlike ordinary attention mechanisms, self-attention centers on capturing relationships within the input sequence itself. In self-attention, the $Q$, $K$, and $V$ in the attention mapping \eqref{eq:attention} are derived from the same input sequence but with distinct representations. 
Denoting the input sequence as $X\in \mathbb{R}^{t \times d'}$, the three matrices $Q$, $K$, and $V$ are computed as weighted transformations of $X$, each achieved using a weight matrix of size $d \times d'$. Specifically, we have $Q = XW_{Q}$, $K = XW_{K}$, and $V = XW_{V}$.
In this perspective, the self-attention mechanism can be represented as a mapping from $\mathbb{R}^{t\times d'}$ to $\mathbb{R}^{t\times d}$, with three weight parameter matrices and an optional masking matrix.
To denote the self-attention of an input sequence $X$ with a tuple of weight parameter matrices $\mathbf{W}$ and a masking matrix $M$, the self-attention $\mathcal{Z}$ is defined as follows:
\bea
\label{eq: attention_f1}
\mathcal{Z}(X; \mathbf{W}, M) = \text{softmax}\left( XW_Q (X W_K)^\top /\sqrt{d} \odot M \right) XW_V,
\eea
where $\mathbf{W} = (W_Q, W_K, W_V)$.
In this paper, we train the weight parameter matrices in $\mathbf{W}$, while keeping the masking $M$ fixed according to the specifications of the multilayer network $\mathcal{G}$ introduced in Section \ref{sec:mscmn}.

\section{Proposed Model} \label{sec: pm}

\begin{figure}[!t]
    \centering
    \includegraphics[width=0.95\linewidth]{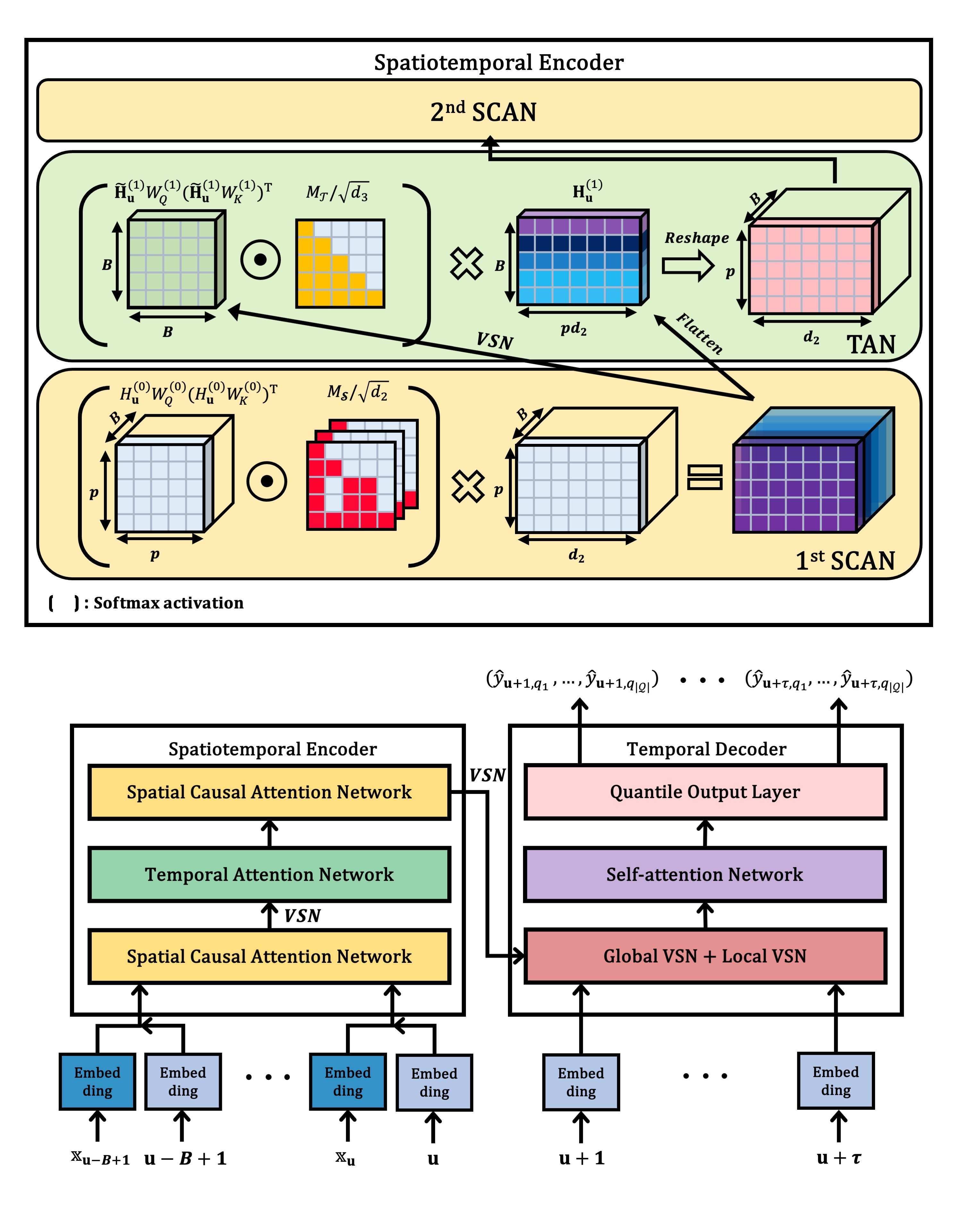}
    \caption{Architecture of the proposed model (bottom) and details of SCAN and TAN in the spatiotemporal encoder (top).}
    \label{fig:model_arch}
\end{figure}

In this section, we present InstaTran (INterpretable SpatioTemporal Attention TRANsformer), an interpretable transformer that integrates spatiotemporal dependencies following the multilayer network structure $\mathcal{G}$ introduced in Section \ref{sec:mscmn}. The overall architecture of the proposed model is displayed in Figure \ref{fig:model_arch}.
All vector notations in this section represent row vectors rather than column vectors.

\subsection{Notations and Model Architecture Overview} \label{sec:oma}

Each observation is indexed by a time feature denoted as $\bu = (u_{1}, u_{2}, u_{3}) \in \mathbb{R}^3$, where $u_{1}$, $u_{2}$, and $u_{3}$ represent the month, day, and hour, respectively. For convenience, we use the notation $\bu+h$ to represent the time feature that is $h$ hours later than $\bu$. Similarly, $\bu-h$ represents the time feature $h$ hours before $\bu$.
The explanatory variable measured at time $\mathbf{u}$ and associated with node $v_{s,t}$ in the multilayer network $\mathcal{G}$ (with $s \in \mathcal{S}$ and $t \in \mathcal{T}$) is denoted as $\mathbf{x}_\bu^{(s,t)} \in \mathbb{R}^{p_s}$, where $p_s$ represents the dimension of the explanatory variable for cluster $C_s$.
A notable feature of our proposed method is the ability for each cluster to have heterogeneous explanatory variables, allowing the dimension $p_s$ to vary depending on the specific cluster $C_s$.
For conciseness, we introduce the notation $\mathbf{x}_\bu$, 
which represents the concatenation of all explanatory variables measured at time $\bu$ from all clusters ($C_s$ with $s \in \mathcal{S}$).
Formally, $\mathbf{x}_\bu = \left(\mathbf{x}^{(1,t)}_\bu, \dots, \mathbf{x}^{(4,t)}_\bu \right) \in \mathbb{R}^p$, where $p = \sum_{s \in \mathcal{S}} p_s$.
Each component in $\bx_\bu$ is denoted by $x_{i,\bu}$, where $i = 1, \dots, p$.
Furthermore, we define a consecutive partition of the set $\{1, \dots, p\}$ denoted by $\mathcal{I} = \{I_1, \dots, I_4\}$, where each $I_s$ represents the set of indices corresponding to $\mathbf{x}^{(s,t)}_\bu$ within $\mathbf{x}_\bu$. From the construction, the cardinality of $I_s$ is equal to $p_s$.

Our model aims to estimate multiple quantiles of the water level as a time series of length $\tau$ measured in hours.
The model takes two time series inputs: one for the explanatory variables of length $B$ and another for the time feature of length $B+\tau$. The input explanatory variable is denoted as $\mathcal{X}_{\bu,B}$, which consists of vectors $\bx_\bu$ for $B$ consecutive hours: $\mathcal{X}_{\bu,B} = (\bx_{\bu-B+1}, \cdots, \bx_\bu)$.  The input time feature is denoted as $\mathcal{U}_{B}^{\tau} = (\bu-B+1, \cdots, \bu, \bu+1, \cdots, \bu+\tau)$, which includes the time feature associated with $\mathcal{X}_{\bu,B}$ as well as time features for the subsequent $\tau$ hours. 
The model outputs the estimation of the time series of water level quantiles for the time period from $\bu+1$ up to $\tau$ hours.
Denoting the set of quantiles of interest by $\mathcal{Q}$,
for each $q \in \mathcal{Q}$, the estimated $q$-quantile of the water level at the time point $\bu$ is denoted as $\hat y_{\bu,q}$, and the output is represented as $\hat \by_{\bu,q}^\tau = (\hat y_{\bu+1,q}, \cdots, \hat y_{\bu+\tau,q})$ for $q \in \mathcal{Q}$.

In our proposed neural network framework, the inputs $\mathcal{X}_{\bu, B}$ and $\mathcal{U}_{B}^{\tau}$ are embedded using self-attention mechanisms associated with spatial masking and temporal causal masking, which are controlled by the masking matrices $M_\mathcal{S}$ and $M_\mathcal{T}$ respectively.
The spatial masking matrix $M_\mathcal{S} \in \mathbb{R}^{p \times p} $ is constructed based on the pre-assumed spatial causal relation presented in Table \ref{tab:causality}.
Denoting the elements at the intersection of the $i$-th row and the $j$-th column of $M_\mathcal{S}$ as $m_{ij}$, they are set according to the following conditions:
\bean
m_{ij} = \begin{cases} 1,  ~ &\mbox{if} ~ i ~ \in I_{s}, \ j \in I_{s'} ~ \mbox{and} ~ v_{s'} \in \mbox{Pa}(v_s) \cup \{ v_s \}  \\
    -\infty, ~ &\mbox{otherwise},
\end{cases}
\eean
where $\text{Pa}(v_s)$ denotes the set of causes of node $v_s$, given by $\text{Pa}(v_s) = \{v_{s'} : (v_{s'},v_s) \in \mathcal{E}\}$. The hour index $t$ can be omitted as the spatial causal relation is assumed to be consistent across all time points.
The temporal causal masking matrix
$M_\mathcal{T}$ equals the decoder attention mask used in a typical language transformer \citep{Vaswani2017AttentionIA}.

In short, our proposed method InstaTran can be represented as follows, with the process function denoted by $F$:
\bean
\hat \by^\tau_{\bu, \mathcal{Q}} = F(\mathcal{X}_{\bu,B}, \mathcal{U}_{B}^{\tau}, M_\mathcal{S}, M_\mathcal{T}, \mathcal{Q}),
\eean
where $\hat \by^\tau_{\bu, \mathcal{Q}}$ represents a tuple $(\hat{\by}^\tau_{\bu, q}, q \in \mathcal{Q})$. 
InstaTran consists of a spatiotemporal encoder and a temporal decoder. The spatiotemporal encoder learns a representation of input variables included in the model, and the temporal decoder produces multiple quantiles of future water levels. 

\subsection{Spatiotemporal encoder}

The spatiotemporal encoder consists of three steps: first, embedding spatial causal relations; second, embedding temporal causal relations; and finally, embedding spatial causal relations once more to enhance their representation.

\subsubsection{Spatially Causal Attention Network} \label{sec:scan}

To embed spatial causal relations, we introduce the Spatially Causal Attention Network (SCAN), a self-attention that embeds a collection of observed variables based on aforementioned spatial causal relations outlined in Table \ref{tab:causality}.
The resulting embedded feature represents the aggregation of information from all sites at a fixed time, with spatial causation across the sites captured by the spatial masking matrix $M_{\mathcal{S}}$ through the use of SCAN.

Before applying SCAN, we construct a matrix $H_\bu^{(0)} \in \mathbb{R}^{p \times d_1}$ to be used as an input for SCAN by embedding, where $d_1$ is the dimension of the output layer. In the following context, $d_\ast$ refers to the output dimension of the neural network layer.
In constructing the $i$-th row of $H_\bu^{(0)}$, the corresponding covariate $x_{i,\bu}$ ($i=1, \cdots, p$) is embedded along with its associated time $\bu$ by a covariate embedding function $g_i: 
\mathbb{R} \rightarrow \mathbb{R}^{d_1}$ and temporal information embedding function $\tilde{g}_j:\mathbb{Z}_j \rightarrow \mathbb{R}^{d_1}$ where $\mathbb{Z}_j ~ (j = 1, 2, 3)$ denote the sets of integers for month, day, and hour, respectively. Specifically, the $i$-th row of $H_\bu^{(0)}$, $h_{i,\bu}^{(0)}$ is constructed as follows:
\bean
h_{i,\bu}^{(0)} = g_i(x_{i,\bu}) + \sum_{j=1}^3 \tilde g_j(u_{j})/3 \in \mathbb{R}^{d_1},
\eean
for $1 \leq i \leq p$.
The feature $\tilde g_j(u_{j})$  plays a dynamic and trainable role in positional encoding within the transformer \citep{Vaswani2017AttentionIA}.

Next, SCAN is applied to $H_\bu^{(0)}$, performing self-attention mapping from $\mathbb{R}^{p \times d_1}$ to $\mathbb{R}^{p \times d_2}$ using a triplet of $d_1 \times d_2$ attention weight matrices ${\bf W}^{(0)} = \left(W^{(0)}_{Q}, W^{(0)}_{K}, W^{(0)}_{V}\right)$, which is defined as follows:
\bea
H^{(1)}_\bu = \mathcal{Z} 
\left(H^{(0)}_\bu; {\bf W}^{(0)}, M_{\mathcal{S}} \right) \in \mathbb{R}^{p \times d_2}. \label{eq: scan1}
\eea
The output of SCAN, denoted as $H^{(1)}_\bu$ in \eqref{eq: scan1}, provides a refined representation of the explanatory variable $\bx_\bu$. By learning the spatial causal relations specified in Assumption \ref{assum2} through SCAN, it gains the capability to address spatial causal relations.
 
\subsubsection{Temporal Attention Network}\label{sec:tan}

We introduce the Temporal Attention Network (TAN), which is a specially designed self-attention that takes the time series of spatial features obtained by SCAN  as its input.  
Specifically, at a given time feature $\bu$, TAN utilizes spatial features ${\bf H}_{\bu'}^{(0)}$ in \eqref{eq: scan1} evaluated at time features $\bu' = \bu-B+1, \cdots, \bu$, forming a time series denoted as ${\bf H}_\bu^{(1)}$ through its construction:
\bea \label{eq:Phi_u}
{\bf H}^{(1)}_{\bu} = \left[\mbox{vec}\left(H^{(1)}_{\bu-B+1}\right)^\top, \dots, \mbox{vec}\left(H^{(1)}_{\bu}\right)^\top \right]^\top \in \mathbb{R}^{B \times pd_2}, 
\eea
where $\mbox{vec}(\cdot)$ denotes a flattening map. Alongside ${\bf H}^{(1)}_{\bu} $ in \eqref{eq:Phi_u}, TAN also utilizes the dimension-reduced representation of ${\bf H}^{(1)}_{\bu}$ in constructing the attention weights. The column size 
is reduced from $p d_2$ to $d_2$ by a variable selection network (VSN, \cite{lim2021temporal}),
and the reduced time series is denoted as $\tilde {\bf H}^{(1)}_{\bu}$:
\bea
\tilde {\bf H}^{(1)}_{\bu} = \left[\mbox{VSN}_1\left(H^{(1)}_{\bu-B+1}\right)^\top , \dots ,  \mbox{VSN}_1\left(H^{(1)}_{\bu}\right)^\top \right]^\top \in \mathbb{R}^{B \times d_2}. \label{eq: Phi}
\eea
In \eqref{eq: Phi}, the $\mbox{VSN}_1(\cdot)$ denotes a VSN layer that compresses separate local information within a given matrix and transforms it into a single vector. 
Consequently, each row of $\tilde{\bf H}_{\bu}^{(1)}$ is  a reduced vector of length ${d_2}$, where the $i$-th row corresponds to $H_{\bu'}^{(1)}$ with $\bu' = \bu-B+i$.
Detailed information of $\mbox{VSN}_1(\cdot)$ is provided in the Appendix. The subscript of $\mbox{VSN}_1(\cdot)$ is employed to differentiate the steps at which the VSN is utilized, given its pervasive usage throughout the entire procedure.

Then, the proposed self-attention TAN is defined as 
\bean 
&& \mbox{TAN}\left( \tilde {\bf H}^{(1)}_{\bu}, {\bf H}_{\bu}^{(1)}; {\bf W}^{(1)}, M_\mathcal{T} \right) = A^{(1)}_\bu {\bf H}_{\bu}^{(1)} \in \mathbb{R}^{B \times pd_2} \label{eq: tan},
\eean
where the attention weight $A^{(1)}_\bu$ is computed as
\bean
A^{(1)}_\bu = \mbox{softmax}\left(\tilde{ \bf H}^{(1)}_{\bu} W^{(1)}_{ Q} \left(\tilde{ \bf H}^{(1)}_{\bu} W_{K}^{(1)}\right)^\top / \sqrt{d_3} \odot M_\mathcal{T} \right).   \eean
The output of TAN is denoted as ${\bf H}_{\bu}^{(2)}$, so that
\bean
{\bf H}_{\bu}^{(2)} = A^{(1)} {\bf H}_{\bu}^{(1)} \in \mathbb{R}^{B \times pd_2}
\eean
Here, ${\bf W}^{(1)} = \left(W^{(1)}_{Q}, W^{(1)}_{K}\right) \in \mathbb{R}^{d_2 \times d_3} \times \mathbb{R}^{d_2 \times d_3}$ represents the pair of trainable weight matrices, and $M_\mathcal{T}$ is the temporal causal mask $M_\mathcal{T}$. The temporal mask $M_\mathcal{T}$ encodes the irreversibility of the temporal features by setting the upper diagonal elements to $-\infty$. Consequently, for $i < j$, the element on the intersection of the $i$-th row and the $j$-th column of $A^{(1)}$ becomes $0$, and the $i$-th row of TAN  is solely composed of the $j$-th row vector of \({\bf H}^{(1)}_{\bu}\). Thereby, TAN adheres to the irreversibility of the temporal features.

All self-attention outputs from this section to Section \ref{sec:td} are indexed by the time feature $\mathbf{u}$, while they are not exclusively constructed from $\mathbf{u}$ alone. Instead, they are formed by aggregating information across time features from $\mathbf{u}-B+1$ to $\mathbf{u}$.

\subsubsection{Strengthening Spatial Causal Relations} \label{sec:ssd}

In employing TAN, the spatial causal relation carried in the input $\tilde {\bf H}^{(1)}_{\bu}$ might be blurred due to the incorporation of VSN, as shown in \eqref{eq: Phi}. As a result, the output of TAN, $\tilde {\bf H}^{(1)}_{\bu}$, could have a weakened representation of spatial causal relations.
To strengthen the spatial causal relations in feature representations, we introduce an additional self-attention step. This step is facilitated by applying SCAN to the outputs of TAN, thereby enhancing the model's ability to capture spatial causal relations.

The procedure is similar to \eqref{eq: scan1} in Section \ref{sec:scan}. We apply self-attention with mask $M_{\mathcal{S}}$ with input ${\bf H}_{\bu}^{(2)}$ in a row-wise manner. Specifically, denoting the row of ${\bf H}_{\bu}^{(2)}$ that corresponds to time feature $\bu'$ ($\bu' = \bu-B+1, \cdots, \bu$) as $h^{(2)}_{\bu'}$, we reshape it into a $p \times d_2$ matrix $H^{(2)}_{\bu'}$, and then feed it to SCAN as follows:
\bea
H^{(3)}_{\bu'} =\mathcal{Z}\left(H^{(2)}_{\bu'}; {\bf W}^{(2)}, M_{\mathcal{S}}\right) = A^{\mathcal{S}}_{\bu'}\left(H^{(2)}_{\bu'}W_V^{(2)}\right)\label{eq: scan2}.
\eea
This approach enables the output ${H}_{\bu'}^{(3)}$ to exhibit reinforced the spatial causal relation over ${\bf H}_{\bu}^{(2)}$, thereby adding spatial causal relation on top of temporal causal relation present in ${\bf H}_{\bu}^{(2)}$ and resulting in more enriched representation learning. The attention weights $A^{\mathcal{S}}_{\bu'}$ in  \eqref{eq: scan2} are utilized as an interpretation measure, as they capture spatial effects in a quantitative manner.

After performing the second SCAN, the final output of the InstaTran's  encoder, denoted as ${\bf H}^{(3)}_{\bu}$,
is computed by utilizing $H^{(3)}_{\bu'}$ in \eqref{eq: scan2}.
Specifically, each row of ${\bf H}_{\bu}^{(3)}$ is constructed by applying VSN on $H^{(3)}_{\bu'}$ for $\bu' \in \{\bu-B+1, \cdots, \bu\}$, similar to \eqref{eq: Phi}:
\bea 
{\bf H}^{(3)}_{\bu} = \left[\mbox{VSN}_2\left(H^{(3)}_{\bu-B+1}\right)^\top, \dots, \mbox{VSN}_2\left(H^{(3)}_{\bu}\right)^\top\right]^\top \in \mathbb{R}^{B \times d_2}. \label{eq: Psi}
\eea
By employing VSN in \eqref{eq: Psi}, the input information is efficiently summarized and transferred to the decoder, while also providing variable selection weights. These weights allow us to assess the importance of variables for forecasting $\hat{\by}^{\tau}_{\bu, \mathcal{Q}}$ \citep{lim2021temporal}.
 The numerical results on interpretation are discussed in Section \ref{sec: interpret}.

\subsection{Temporal Decoder} \label{sec:td}

We propose an architecture leveraging global and local context vectors, constructed via the feedforward network (FFN) layer, specifically  VSN. This approach, inspired by \cite{wen2017multi}, enables the simultaneous prediction of future events up to $\gamma$ time points. In contrast, competitive methods like TFT recursively forecast by relying on previous time point predictions.
our method incorporates two VSN layers within the decoder: one for global and another for local context.

The global VSN summarizes ${\bf H}_{\bu}^{(3)}$ and constructs a global context vector $h^{(4)}_{\bu} \in  \mathbb{R}^{d_2}$ as follows:  
\bea \label{eq: global}
h^{(4)}_{\bu} = \mbox{VSN}_3\left({\bf H}_{\bu}^{(3)}\right).
\eea
The local VSN generates the local context vector $\tilde g_{\bu}$ by utilizing the temporal features embedded during the encoding step discussed in Section \ref{sec:scan}. As such, $\tilde g_{\bu}$ captures a sense of locality, and it is constructed as follows:
\bea
\tilde g_{\bu} = \mbox{VSN}_4\left(\left[\tilde g_1(u_1)^\top, \tilde g_2(u_2)^\top, \tilde g_3(u_3)^\top\right]^\top W^{(3)} + b^{(3)}\right) \in \mathbb{R}^{d_2}, \label{eq: vsn_local}
\eea
where $W^{(3)} \in \mathbb{R}^{d_1 \times d_2}$ and $b^{(3)} \in \mathbb{R}^{d_2}$ denote trainable weight and bias vector, respectively. 
The two outputs of VSN, shown in \eqref{eq: global} and \eqref{eq: vsn_local}, form
a pooled context vector ${\bf Z}_\bu$ as follows:
\bean
{\bf Z}_\bu = \left[z_{\bu+1}^\top, \cdots, z_{\bu+\tau}^\top \right]^\top \in \mathbb{R}^{\tau \times d_2},
\eean
where $z_{\bu+k} = h_{\bu}^{(4)} + \tilde g_{\bu+k}$ for $k = 1, \cdots, \tau$.  The pooled context vector ${\bf Z}_\bu $  pertains to the temporal features starting from the time point $\bu + 1$ and extending into the subsequent $\tau$ steps.

Subsequently, the encoder's output in \eqref{eq: scan2} and the pooled context vector are concatenated to form ${\bf H}_\bu^{(4)}$, juxtaposing the evaluated features from  $B$ steps backward and $\tau$ steps forward, starting from time point $\bu$. Specifically, ${\bf H}_\bu^{(4)}$ is constructed as follows: 
\bean
{\bf H}_\bu^{(4)} = \left[{\bf H}_\bu^{(3)\top}, {\bf Z}_\bu^{\top}\right]^\top \in \mathbb{R}^{(B+\tau) \times d_2}.
\eean
This composite matrix is then directed into a self-attention network, yielding the augmented temporal feature representation denoted as ${\bf H}_\bu^{(5)}$:
\bea \label{eq:last_attention}
{\bf H}_\bu^{(5)} = \mathcal{Z}\left({\bf H}^{(4)}_{\bu}; {\bf W}^{(4)}, M_\mathcal{T}\right) = A^{\mathcal{T}}_{\bu}\left( {\bf H}^{(4)}_{\bu} W^{(4)}_{V}\right) \in \mathbb{R}^{(B+\tau) \times d_3},
\eea
where ${\bf W}^{(4)}  = \left(W^{(4)}_{Q}, W^{(4)}_{K}, W^{(4)}_{V}\right)$.
The output ${\bf H}_\bu^{(5)}$ in \eqref{eq:last_attention} is the final output of the decoder layer of InstaTran.
Temporal importance can be measured by assessing the attention weights $A^{\mathcal{T}}_{\bu}$ obtained from the final self-attention layer, as shown in \eqref{eq:last_attention}. Through an examination of these attention weights, we discern the past time point to which our model allocates its focus. This analysis further enables us to assess the alignment of these attention patterns with the predefined assumptions fed to the model.

The forecasting of water level quantiles $\hat{y}_{\bu+k, q} \in \mathbb{R}$, where $k = 1, \dots, \tau$, which is the targeted output of our proposed model, is achieved through the utilization of the FFN layer on ${\bf H}_\bu^{(5)}$ as follows:
\bean
\hat{y}_{\bu+k,q} &=& \left({\bf H}_\bu^{(5)} \right)_{B+k, :} W_q^{(5)} + b_q^{(5)}, ~ q \in \mathcal{Q},
\eean
where $W_{q}^{(5)} \in \mathbb{R}^{d_3 \times 1}, b_q^{(5)} \in \mathbb{R}^{1}$ are trainable parameters, and $\left({\bf H}_\bu^{(5)}\right)_{B+k, :}$ denotes the $(B+k)$-th row of ${\bf H}_\bu^{(5)}$, $k=1, \dots, \tau$. 
A distinctive attribute of our proposed approach is its direct forecasting of $\hat{\by}_{\bu,q}^\tau$, which is a tuple of $\tau$ consecutive quantiles. This stands in contrast to other models, which employ a recursive forecasting process by building upon the preceding forecasting for the subsequent forecast via RNN layers. Consequently, the proposed model maintains a simple and efficient architecture when compared to TFT. Previous studies suggest that decoders designed for direct forecasting often enhance performance by mitigating the accumulation of errors, which in turn can prevent biased predictions \citep{Chevillon2006DirectME, Taieb2016ABA, wen2017multi}. Our empirical analysis of real-world data further supports this observation, demonstrating superior outcomes compared to TFT, as detailed in Section  \ref{sec: abl}.

\subsection{Loss functions} \label{sec: loss}

In the training phase, InstaTran is instructed to minimize the composite quantile loss (CQL), which comprises a collection of quantile losses. The quantile loss is defined as follows:
\bea \label{eq: ql}
\mbox{QL}(y, \hat{y}; q) = \left(q - \mathbb{I}_{\{y < \hat{y}\}}(y)\right)\left(y-\hat{y}\right),
\eea
where $ \mathbb{I}_{\mathcal{A}}(a)$ returns 1 for $a \in \mathcal{A}$ and 0, otherwise. The CQL is defined as follows:
\bean
\mbox{CQL}(\mathbb{W}; \bU, \mathcal{Q}, \tau) = \sum_{\bu \in \bU} \sum_{q \in \mathcal{Q}}\sum_{k=1}^\tau \mbox{QL}\left(y_{\bu+k}, \hat{y}_{\bu+k, q};q\right),
\eean
where $\mathbb{W}$ denotes the entire weight and bias parameters, and $\bU$ is the set of time points in the training dataset.

\section{Experiments}

We evaluate the effectiveness of the proposed model through an analysis of its performance in probabilistic forecasting and interpretability. This evaluation is conducted using real-world data of the Han River water level dataset, which is discussed in Section \ref{sec:data}.
For comparative analysis, we include nine benchmark models: 
ETS (Error, Trend, and Seasonality), ARIMA \citep{box1994time}, Theta \citep{assimakopoulos2000theta}, LightGBM \citep{ke2017lightgbm} with Fourier terms with daily period and four components: $\{\cos \frac{2\pi n t}{24}, \sin \frac{2\pi n t}{24}\}_{n=1}^4$ at time point $t \in \{\bu - B + 1, \dots, \bu \}$, STA-LSTM \citep{Ding2020InterpretableSA}, HSDSTM \citep{Deng2022ASG}, DeepAR \citep{salinas2020deepar}, MQ-RNN \citep{wen2017multi}, and TFT \citep{lim2021temporal}.
ETS, ARIMA, and Theta are statistical models, LightGBM\footnote{The LightGBM API currently does not support composite quantile loss. Therefore, the LightGBM model is individually fitted to each quantile loss.} is a tree-based model, and HSDSTM, DeepAR, MQ-RNN, and TFT are deep learning-based models.
Among deep learning-based models, STA-LSTM and HSDSTM capture domain-specific information by utilizing spatiotemporal structure.
STA-LSTM leverages both LSTM and attention mechanisms to capture complex spatial and temporal dependencies. HSDSTM utilizes a temporal convolution network (TCN) for a long-term dependency and exploits spatial dependencies from graph-structured data with GNN.
Among the investigated models, TFT exhibits the highest complexity, with a total of 99,497 parameters tailored to our specific problem.
The parameter counts for the remaining models are as follows: InstaTran - 77,047, DeepAR - 45,614, HSDSTM - 42,033, STA-LSTM - 15,933, and MQ-RNN - 4,099, listed in descending order.

The dataset is split into two segments: the training dataset from  2016 to 2020 and the test dataset in 2021. The hyperparameters of all models are selected through cross-validation. Detailed hyperparameter settings are provided in the Appendix. 
All the considered models make forecasting for the water level (WL) of $B_0$,  over a 12-hour period, utilizing data from the preceding 48 hours of data (i.e., $\tau=12$ and $B=48$). 
To encompass a spectrum from regular conditions to high-impact events like flooding, the targeted quantile levels are set to $\mathcal{Q}=\{0.1, 0.5, 0.7, 0.9\}$ during the training stage. The evaluation measure values presented for the test data correspond to the quantiles $\{0.5, 0.7, 0.9\}$. The experiments were conducted using \textsf{PyTorch} and \textsf{sktime} on an NVIDIA GeForce RTX 3090, and the source code is publicly accessible at \url{https://github.com/chulhongsung/InstaTran}.

To assess performance in probabilistic forecasting, we employ two evaluation measures: the quantile loss, discussed in \eqref{eq: ql}, and the calibration metric $q$-Rate \citep{chen2012forecasting, wen2017multi}, which are as follows:
\bea
q\mbox{-level QL} &=& \sum_{\bu \in  \bU'}\sum_{k=1}^\tau \mbox{QL}\left(y_{\bu+k}, \hat{y}_{\bu+k, q};q\right) \label{def: ql_loss},\\
q\mbox{-Rate} &=& \sum_{\bu \in  \bU'} \sum_{k=1}^\tau \frac{\mathbb{I}_{\{y_{\bu+k} < \hat{y}_{\bu+k, q}\}}(y_{\bu+k})}{|\bU'|\tau}, \label{def: qrate}
\eea
where $\bU'$ denotes the set of time points corresponding to the test dataset. 
The $q$-Rate is defined as the proportion of observations that fall below the forecasted value of the $q$-th quantile. When the $q$-Rate closely aligns with the target quantile value $q$, it indicates strong performance.

\subsection{Ablation studies of InstaTran} \label{sec: abl}
\begin{figure}[!t]
    \centering
    \subfigure[Dry day with $M_{\mathcal{S}}$]{\includegraphics[width=0.46\linewidth, height=4.8cm]{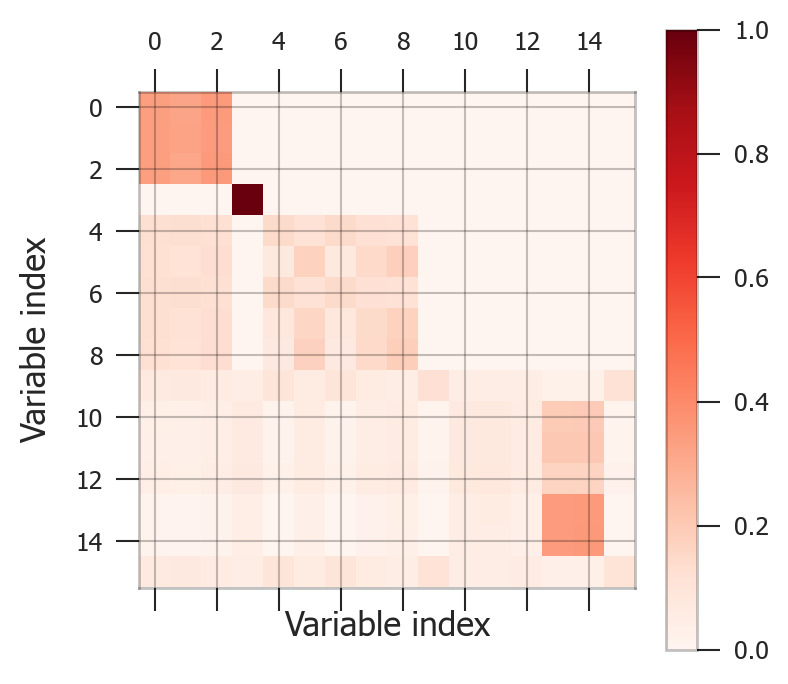}}
    \subfigure[Dry day without $M_{\mathcal{S}}$]{\includegraphics[width=0.46\linewidth]{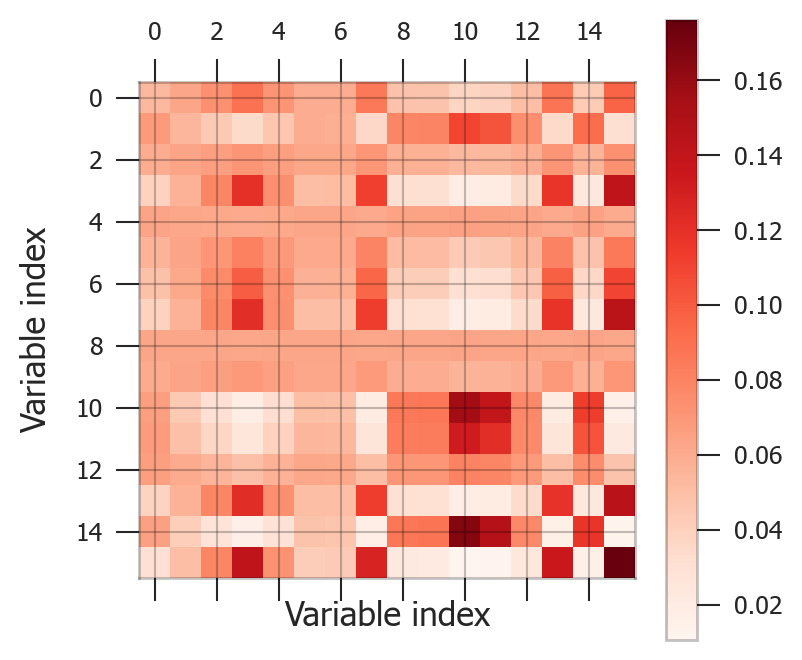}}
    \subfigure[Rainy day with $M_{\mathcal{S}}$]{\includegraphics[width=0.46\linewidth, height=5cm]{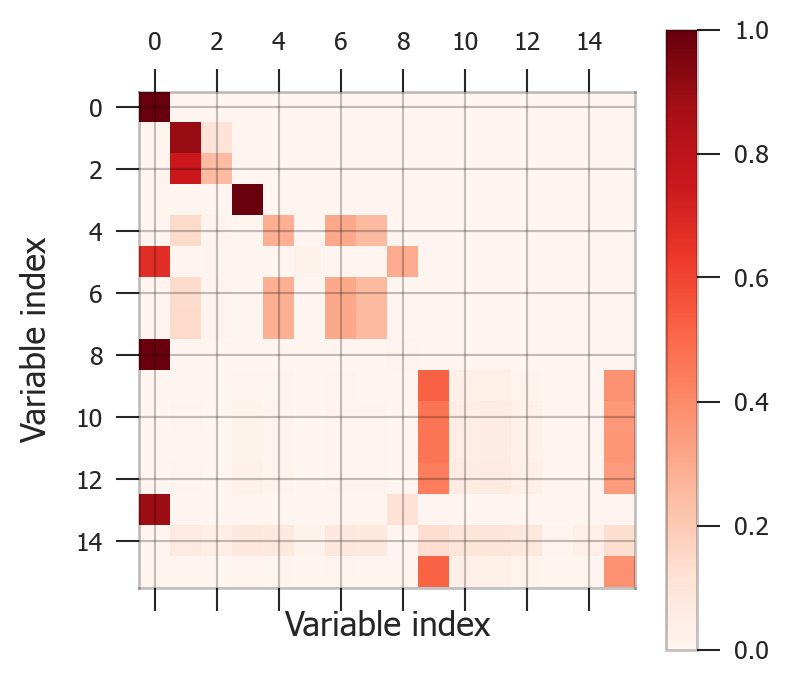}}
    \subfigure[Rainy day without $M_{\mathcal{S}}$]{\includegraphics[width=0.46\linewidth, height=5cm]{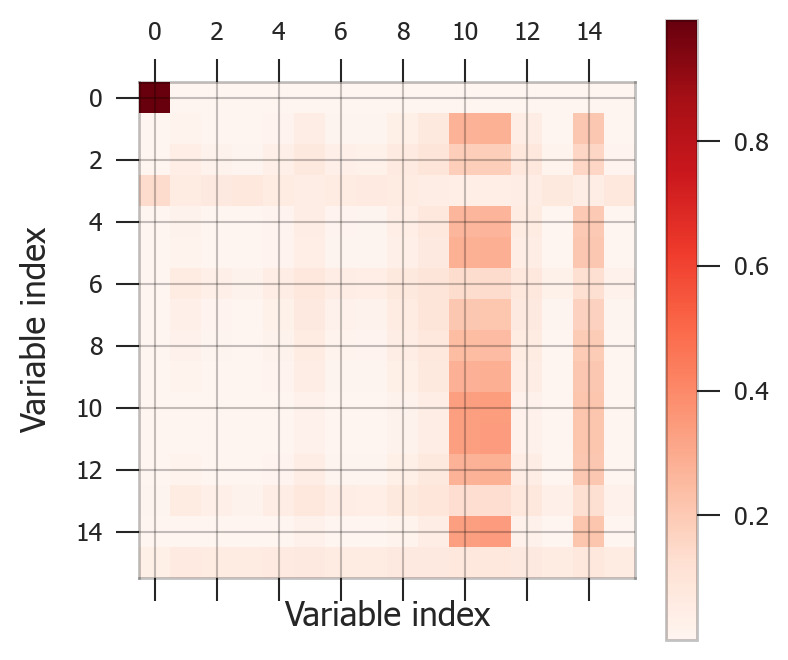}}
    \caption{Heatmaps of attention weights $A^{\mathcal{S}}_{\bu'}$, disussed in \eqref{eq: scan2}. Plots (a) and (b) on the top row correspond to a dry day, and plots (c) and (d) on the bottom row correspond to a rainy day. The variables corresponding to the indices of both the x-axis and y-axis are  as follows: $0:P_1,~1:P_2, ~2:P_3,~3:\mbox{WL}~(B_4),~4:\mbox{WL}~(D),~5:\mbox{IF}~(D),~6:\mbox{STR}~(D),~7:\mbox{JUS}~(D),~8:\mbox{OF}~(D),~9:\mbox{WL}~(B_1),~10:\mbox{FL}~(B_1),~11:\mbox{WL}~(B_0),~12:\mbox{WL}~(B_2),~13:\mbox{FL}~(B_2),~14:\mbox{WL}~(B_3),~15:\mbox{FL}~(B_3)$.}
    \label{fig: saw}
\end{figure}

Before presenting the comparison of benchmark models, we explore the results of our ablation studies to highlight the strengths of our proposed model. Initially, we present the efficacy of the proposed spatiotemporal encoders, namely SCAN and TAN in InstaTran. Subsequently, we demonstrate the effectiveness of the overall model architecture.

To showcase the impact of the proposed encoders, we provide heatmaps illustrating attention weights $A^{\mathcal{S}}_{\bu'}$ in Figure \ref{fig: saw}, that are evaluated with and without the incorporation of $M_\mathcal{S}$ in \eqref{eq: scan2}. 
These evaluations account for two scenarios: rainy days and dry days.
In the heatmap matrix depicted in Figure \ref{fig: saw}, the colors within the cell at the intersection of the $i$-th row index and $j$-th column index representing the importance of the $j$-th variables contributing to the output of the $i$-th feature.
For example, the cells on the column indexed by $0$ in Figure \ref{fig: saw} (c) exhibit darker shades in the rows indexed by $5$ and $8$. This implies the strong influence of the $0$-th features of $H_{\bu'}^{(2)}$ on the composition of $5$-th and $8$-th features of $H_{\bu'}^{(3)}$, aligning with our spatial causal relation $v_1 \rightarrow v_3$ in Assumption \ref{assum2}. Furthermore, the results obtained with the mask demonstrate plausible outcomes, as they clearly differentiate between rainy and dry days, capturing the impact of rainfall in accordance with our expectations.
Conversely, when $M_{\mathcal{S}}$ is not applied, the attention weight $A^{\mathcal{S}}_{\bu'}$ does not adhere to the presumed spatial causal relation in Table \ref{tab:causality}, nor does it yield a meaningful interpretation. The analysis results of domain-specific methods, STA-LSTM and HSDSTM, are presented in the Appendix. Both models exhibit limitations in capturing dynamic spatial dependencies. This is primarily due to their reliance on a simplistic data structure assumption, such as data homogeneity. 
While these methods assume identical feature presence across all sites, our dataset consists of flexible features.

\begin{table}[t]
    \centering
    \caption{Performance comparison among variants of InstaTran across the four scenarios. The most favorable outcomes are indicated in bold.}
    \begin{adjustbox}{width=\textwidth}
    \begin{tabular}{c c c c c c} 
    \Xhline{1.5pt}  
    Metric &  $q$ & Parallel Attentions & Without $M_{\mathcal{S}}$ & With TFT decoder & InstaTran \\ \hline 
    \multirow{3}{*}{average $q$\mbox{-level QL}} & 0.9 & 0.0034 & 0.0025 & 0.0031  & \textbf{0.0021}  \\
    & 0.7 & 0.0072 & 0.0045 & 0.0051 & \textbf{0.0036} \\
    & 0.5 & 0.0086 & 0.0048 & 0.0059 & \textbf{0.0040} \\
    \multirow{3}{*}{$q$-Rate ($|q - q\mbox{-}\mbox{Rate}|$)} & 0.9 & 0.936 (0.036) & 0.946 (0.046) & 0.798 (0.102) & \textbf{0.924 (0.024)} \\
    & 0.7 & 0.894 (0.194)& 0.838 (0.138) & \textbf{0.638 (0.062)} & 0.796 (0.096) \\
    & 0.5 & 0.823 (0.323) & 0.666 (0.166) &  \textbf{0.623 (0.123)} & 0.647 (0.147) \\
    \Xhline{1.5pt}   
    \end{tabular}
    \end{adjustbox}
    \label{tab:result_ablation}
\end{table}
  
Next, we analyze the forecasting performances of the proposed spatiotemporal encoder and temporal decoder in four different scenarios.
In the first scenario, temporal and spatial attentions are employed in a parallel fashion, allowing both SCAN and TAN to simultaneously receive their respective hidden features. 
This contrasts the sequential approach we proposed, in which SCAN and TAN are applied successively.
This parallel arrangement enables the independent learning of features.
The second scenario investigates the SCAN method without utilizing the masking $M_\mathcal{S}$, aiming to shed light on the role of masking.
In the third scenario, we employ the TFT decoder instead of the temporal decoder outlined in Section \ref{sec:td} of our proposed architecture.
For comparison, the fourth involves the proposed InstaTran in its original form.
Table \ref{tab:result_ablation} presents the favorable forecasting performance achieved by the proposed architecture. In comparison to other explored scenarios, the original InstaTran exhibits enhanced forecasting accuracy across multiple quantile levels. Notably, it also demonstrates superior performance, particularly at the high quantile level of $0.9$.

\subsection{Interpretation of model prediction based on variable importance} \label{sec: interpret}
In this section, we provide interpretations of the prediction results from the InstaTran, TFT, and LightGBM models. 
For InstaTran and TFT, interpretations are demonstrated by evaluating variable importance via the VSN layer. For LightGBM, variable importance is assessed based on the number of splitting nodes for specific variables.

\subsubsection{Variable importance analysis in InstaTran and TFT}
We evaluate the variable importance obtained from InstaTran and compare it with the results from TFT, which serves as one of our benchmark models.
For both InstaTran and TFT, the variable importance is determined through the VSN layer at each time feature $\bu$. 
In InstaTran, the variable importance is established as the weights of the final VSN layer in the encoding step in \eqref{eq: Psi}. On the other hand, in TFT, the variable importance is defined by the weights of the VSN layer located at the input layer.
The weights of VSN are positive and sum up to 1. Thereby, they can be interpreted as the contributions of the variables towards the output. 

\begin{table}[t]
    \centering
    \caption{Descriptive statistics of variable importance. The most significant variables are marked in bold.}
    \begin{adjustbox}{width=\textwidth}
    \begin{tabular}{c c c c c c c c c c} 
    \Xhline{1.5pt}  
         &  & \multicolumn{4}{c}{TFT} & \multicolumn{4}{c}{InstaTran} \\
        \cmidrule(lr){3-6} \cmidrule(lr){7-10}
        Node & Variable & Mean (Std) & 0.1            & 0.5            & 0.9           & Mean (Std)    & 0.1            & 0.5            & 0.9   \\ \hline 
   \multirow{3}{*}{$v_1$} & \text{$P_1$} & 0.031 (0.007) & 0.023          & 0.031          & 0.042      & 0.092 (0.099)   & 0.006          & 0.062          & 0.249 \\
    & \text{$P_2$} & 0.021 (0.014) & 0.008          & 0.017          & 0.030      & 0.042 (0.030)   & 0.010          & 0.038          & 0.079 \\
    & \text{$P_3$} & 0.034 (0.015) & 0.019          & 0.031          & 0.053      & 0.061 (0.028)   & 0.034          & 0.053          & 0.104 \\
    \hline 
    $v_2$ & \text{WL ($B_4$)} & 0.131 (0.016) & 0.110          & 0.132          & 0.153    & \textbf{0.179 (0.070)}   & \textbf{0.058}          & \textbf{0.201} & \textbf{0.251} \\
    \hline
    \multirow{5}{*}{$v_3$}& \text{WL ($D$)}  & 0.018 (0.015) & 0.006          & 0.013          & 0.040      & 0.088 (0.075)   & 0.009          & 0.063          & 0.198 \\
    & \text{IF ($D$)}  & 0.074 (0.021) & 0.005          & 0.070          & 0.101      & 0.074 (0.038)   & 0.034          & 0.066          & 0.124 \\
    & \text{STR ($D$)} & 0.018 (0.007) & 0.010          & 0.017          & 0.027      & 0.006 (0.005)   & 0.003          & 0.005          & 0.011 \\
    & \text{JUS ($D$)} & 0.012 (0.011) & 0.003          & 0.008          & 0.027      & 0.057 (0.023)   & 0.034          & 0.053          & 0.083 \\
    & \text{OF ($D$)}  & 0.096 (0.013) & 0.080          & 0.095          & 0.111      & 0.048 (0.025)   & 0.023          & 0.040          & 0.088 \\
    \hline
    \multirow{7}{*}{$v_4$} & \text{WL ($B_0$)} & \textbf{0.249 (0.077)} & \textbf{0.148} & \textbf{0.253} & \textbf{0.348} & 0.084 (0.048) & 0.045 & 0.064        & 0.160 \\
    & \text{WL ($B_1$)} & 0.040 (0.014) & 0.025    & 0.037          & 0.059      & 0.091 (0.023)   & 0.055          & 0.097          & 0.114 \\
    & \text{FL ($B_1$)} & 0.015 (0.007) & 0.008          & 0.013          & 0.023      & 0.022 (0.015)   & 0.008          & 0.017          & 0.046 \\
    & \text{WL ($B_2$)} & 0.019 (0.004) & 0.015          & 0.019          & 0.025      & 0.054 (0.046)   & 0.008          & 0.038          & 0.117 \\
    & \text{FL ($B_2$)} & 0.019 (0.012) & 0.009          & 0.016          & 0.034      & 0.040 (0.081)   & 0.006          & 0.014          & 0.083 \\
    & \text{WL ($B_3$)} & 0.082 (0.024) & 0.058          & 0.077          & 0.114      & 0.043 (0.046)   & 0.011          & 0.020          & 0.114\\
    & \text{FL ($B_3$)} & 0.016 (0.010) & 0.007          & 0.014          & 0.028      & 0.014 (0.040)   & 0.050          & 0.004          & 0.026 \\
    \Xhline{1.5pt}   
    \end{tabular}
    \end{adjustbox}
    \label{tab:result_vsw}
\end{table}

Table \ref{tab:result_vsw} presents the mean, standard deviation, and quantiles of the variable importance of each variable
obtained through the VSN layers, evaluated across the time points in the test set.
One notable feature in the table is that the median of the variable importance for all variables corresponding to $v_1$ in InstaTran is at least 70\% higher than those in TFT. This observation supports the notion that InstaTran effectively adheres to Assumption \ref{assum2}.
In Assumption \ref{assum2}, $v_1$ exerts influence on the effect node ($v_4$) through two distinct paths: $v_1 \rightarrow v_4$ and $v_1 \rightarrow v_3 \rightarrow v_4$, while all other cause nodes exert only unidirectional influence on $v_4$.
Consequently, $v_1$ assumes a more important role when Assumption \ref{assum2} is observed.
On the other hand, TFT seems unable to capture the underlying spatial causal relation. One piece of evidence is the TFT's assignment of an unusually high weight to the variable corresponding to water level $B_0$. Its median weight of 0.253 is more than three times larger than the second largest median weight of 0.077.
Considering that the water level variable $B_0$ itself serves as the target variable to be forecasted, TFT's prioritization of it may cause TFT to behave more like an autoregressive model, overlooking the incorporation of spatial context in the forecasting procedure. Consequently, this could pose challenges in interpreting and drawing proper inferences from the obtained results.

The standard deviations presented in Table \ref{tab:result_vsw} capture the variability of the variable importance values within the VSN layer across different time points. In the majority of cases, InstaTran exhibits higher standard deviations than TFT.
This observation supports a heightened sensitivity to contextual factors that impact the water level fluctuations of the Han River. Also, it's noticeable that in the case of InstaTran, the mean tends to be higher than the median.
This implies a right-skewed distribution, indicating the presence of heightened variable importance values that typically correspond to rare events such as intense rainfall.

\begin{figure}[!t]
    \centering
    \subfigure[Variable importance of $P_1$.]{\includegraphics[width=0.46\linewidth]{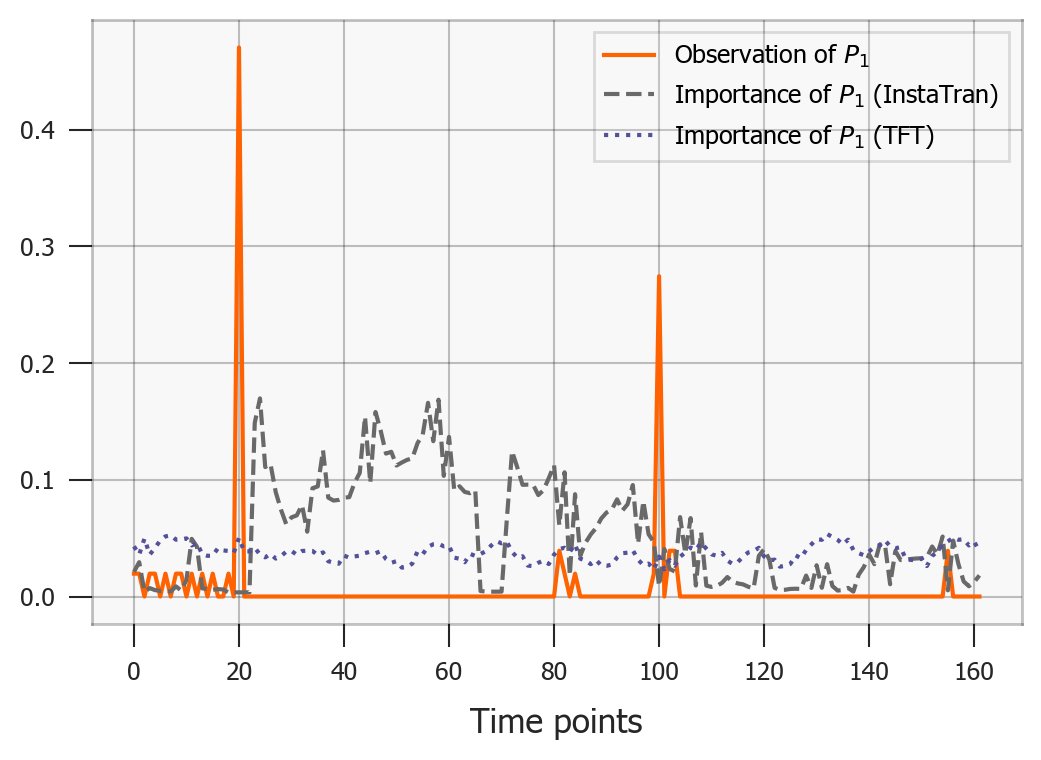}}
    \subfigure[Variable importance of OF$(D)$.]{\includegraphics[width=0.46\linewidth]{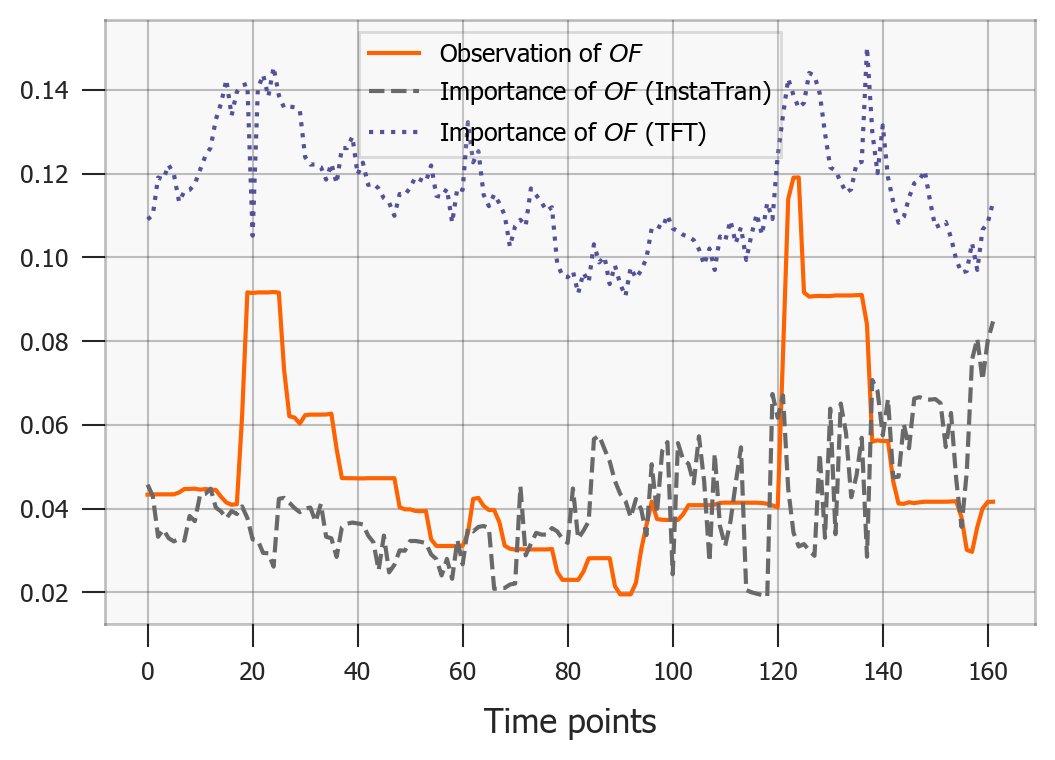}}
    \caption{Variable importances and observations of $P_1$ and OF $(D)$. The solid line denotes the observations of variables. The dashed and dotted lines denote the evaluated variable importances obtained from InstaTran and TFT, respectively.}
    \label{fig: importance}
\end{figure}

Figures \ref{fig: importance} (a) and (b) showcase the enhanced interpretability of InstaTran in comparison to TFT.
These figures display the observations of precipitation $P_1$ and flow discharge OF ($D$), along with their corresponding variable importance of both InstaTran and TFT. 
Notably, at $t=20$, the figures offer a meaningful illustrative example. In Figure \ref{fig: importance} (a), there is a distinct peak in precipitation at this time point, promptly followed by a corresponding surge in flow discharge as displayed in Figure \ref{fig: importance} (b).
Around the same time point, the variable importances obtained from InstaTran and TFT show contrasting patterns. 
In Figure \ref{fig: importance} (a), the variable importance obtained from InstaTran, represented by the dashed line, surges significantly after $t=20$ for a specific duration, while the importance of OF remains consistent. In contrast, the variable importances indicated by TFT demonstrate an inverse trend, where the importance of $P_1$ remains stable while that of OF rises at $t=20$. 
This signifies that InstaTran aligns with our assumption, highlighting the importance of precipitation rather than water discharge in explaining the elevated water levels of the Han River. This supports the common understanding that when intense rainfall leads to a prompt increase in water discharge, the primary driver behind the heightened water level in the Han River is the intense rainfall, rather than the discharge of water.

Another intriguing example is presented in the figures around $t=100$. In Figure \ref{fig: importance} (a), a notable rainfall is observed at $t=100$, followed by a delayed increase in water discharge at $t=120$ as depicted in Figure \ref{fig: importance} (b).
In this instance, InstaTran refrains from assigning greater variable importance to $P_1$ after $t=100$. Instead, InstaTran exhibits a heightened evaluation of the variable importance of $D$. This illustrates that InstaTran does not consistently assign elevated variable importance to $P_1$ after every rain event. Instead, it demonstrates an ability to capture contexts. Specifically, InstaTran encapsulates that the primary cause of water level elevation in this instance is attributed to water discharge.
These two examples exhibit InstaTran's ability to assess the variable importance by simultaneously incorporating both spatial and temporal causal relations. 

\subsubsection{Variable importance analysis in LightGBM}
We compute variable importance in LightGBM by averaging the importance scores across all tree estimators for a given quantile. In each tree, the variable importance is quantified by the number of splitting nodes involving that variable.
Table \ref{tab:fi_lgb} presents the variable importance analysis results of LightGBM. We report the top five variables with the highest importance scores, along with their corresponding time points. Table \ref{tab:fi_lgb} presents that, unlike InstaTran and TFT, LightGBMs consistently consider OF$(D)$ as the most important variable at time point $\bu$ across all quantile levels. Interestingly, a roughly 12-hour pattern is observed (e.g., high scores for WL$(B_0)$ at $\bu$ and WL$(B_0)$ at $\bu -13$ in $q=0.5$, and high scores for WL$(B_4)$ at $\bu$ and WL$(B_4)$ at $\bu-16$ in $q=0.7$). This observation aligns with the temporal patterns found in InstaTran and TFT, which are discussed in Section 4.3.
Additionally, LightGBM tends to prioritize the historical information of the target variable, the water level of $B_0$, over the main causative factor, the water level of $B_4$, as observed in TFT. This characteristic clearly distinguishes LightGBM and TFT from InstaTran.


\begin{table}[h]
    \centering
    \caption{Top 5 variables with the highest variable importance of LightGBM within each quantile level. The value in the parenthesis presents the average splitting numbers in the estimators.}
    \begin{adjustbox}{width=\textwidth}
    \begin{tabular}{c ccc} 
    \Xhline{1.5pt}  
     & \multicolumn{3}{c}{Quantile level (q)}\\
      \cmidrule(lr){2-4}
    Rank & 0.5 & 0.7 & 0.9 \\ 
    1 & OF$(D)$ at time point $\bu$ (11.42) & OF$(D)$ at time point $\bu$ (10.92) & OF$(D)$ at time point $\bu$ (6.83) \\
    2 & WL$(B_0)$ at time point $(\bu - 13)$ (8.08) &  WL$(B_0)$ at time point $(\bu - 13)$ (7.25) & WL$(B_0)$ at time point $\bu - 13$ (5)\\
    3 & WL$(B_0)$ at time point $\bu$ (7.92) & WL$(B_0)$ at time point $\bu$ (5.92) & WL$(B_0)$ at time point $\bu$ (4.67)\\
    4 & IF$(D)$ at time point $\bu$ (4.67) &  WL$(B_4)$ at time point $(\bu - 16)$ (3.67) & WL$(B_4)$ at time point $\bu-17$ (3.25) \\
    5 & WL$(B_4)$ at time point $\bu$ (4.33) &  WL$(B_4)$ at time point $\bu$ (3.42) & WL$(B_0)$ at time point $\bu-12$ (3.08)\\
    \Xhline{1.5pt}   
    \end{tabular}
    \end{adjustbox}
    \label{tab:fi_lgb}
\end{table}

\subsection{Temporal patterns}
In this section, we explore the temporal weights obtained from InstaTran, comparing them with those from TFT.
For InstaTran, we focus on evaluating the attention weight of  $A^{\mathcal{T}}_\bu$ 
in \eqref{eq:last_attention}.  This attention weight matrix is associated with the self-attention within the decoding layer. 
Similarly, for TFT, we evaluate the attention weights within the temporal self-attention layer.

The attention weight $A^{\mathcal{T}}_\bu$ captures the strength of association between predictor variables at previous time points and future prediction time points,  centered around the input time point $\bu$. This capability arises from its construction, where the corresponding self-attention output ${\mathbf H}_\bu^{(5)}$ in \eqref{eq:last_attention} concatenates past observations and available future measurements.
Denoting the element at the intersection of the $(k+48)$-th row and $(t+48)$-th column of $A^{\mathcal{T}}_\bu$ as $(A^{\mathcal{T}}_\bu)_{kt}$ (where $k = 1, \dots, 12$ and $t = -47, \dots, -1, 0, 1, \dots, 12$), this represents the weight that signifies the impact of the observation at the time point $(\bu + t)$ on the prediction at the future time point $(\bu + k)$. Hence, we examine $(A^{\mathcal{T}}_\bu)_{kt}$ to demonstrate the importance of feature variables within the previous hours on future predictions. Specifically, we evaluate the $50$th percentile of $(A^{\mathcal{T}}_\bu)_{kt}$ among the entire period $\bu \in {\mathbf U'}$ corresponding to the test set, and denote it as $w_{50}^k(t)$.

\begin{figure}[!t]
    \centering
    \subfigure[$w_{50}^{k}(t), k \in \{1, \dots, 6\}$ of InstaTran]{\includegraphics[width=0.48\linewidth]{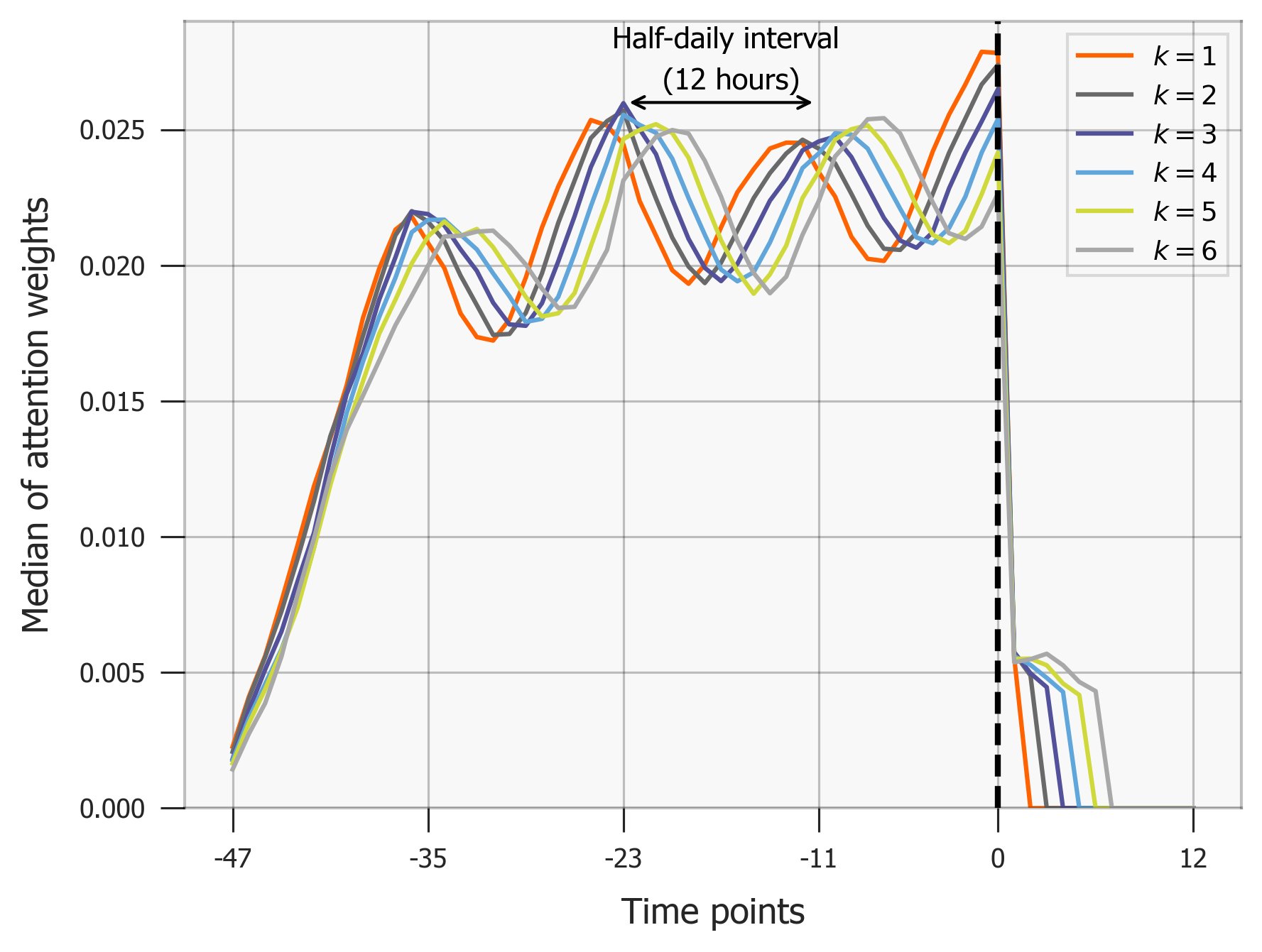}}
    \subfigure[$w_{50}^{k}(t), k \in \{7, \dots, 12\}$ of InstaTran]{\includegraphics[width=0.48\linewidth]{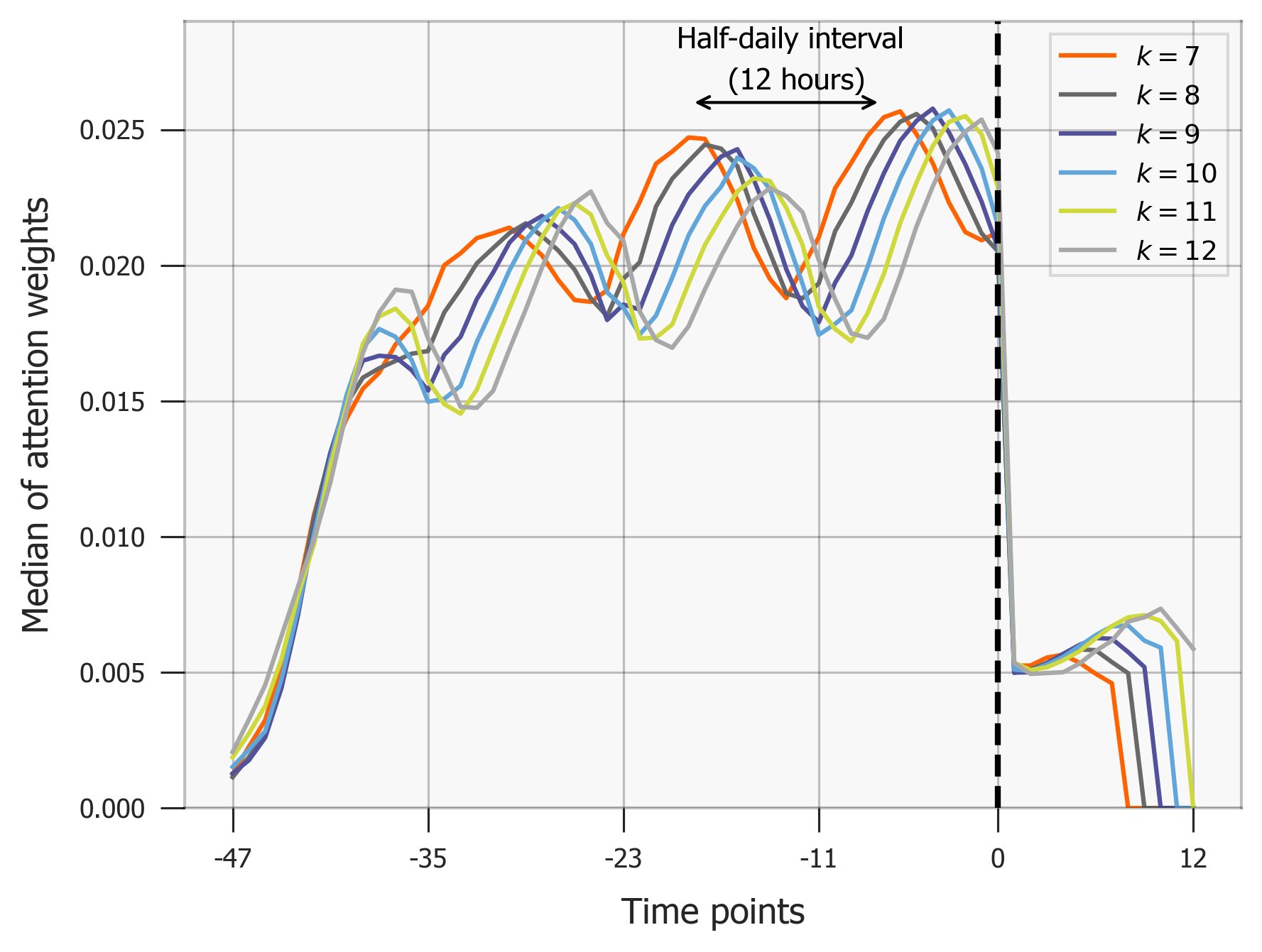}}
    \subfigure[$w_{50}^{k}(t), k \in \{1, \dots, 6\}$ of TFT]{\includegraphics[width=0.48\linewidth]{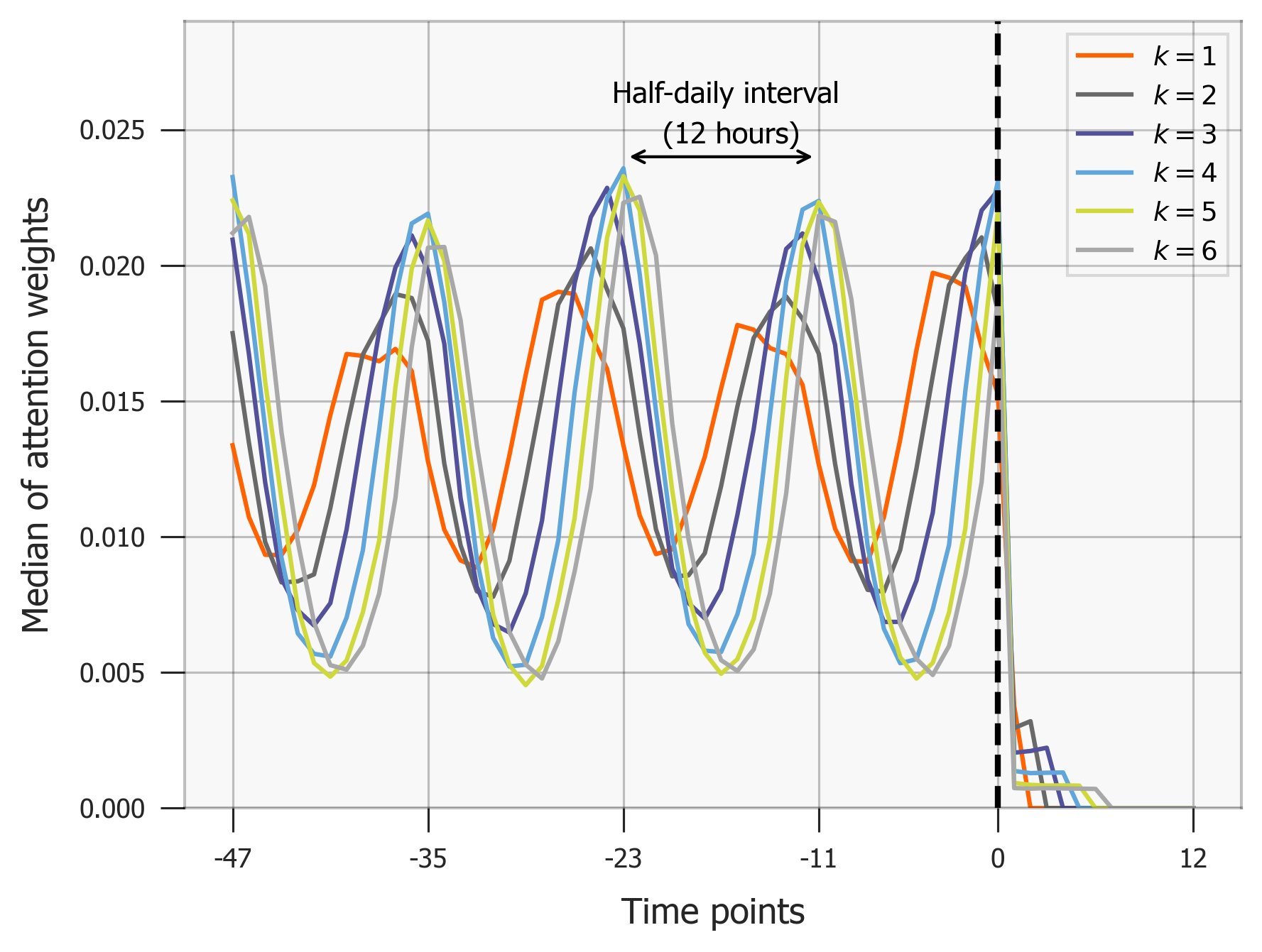}}
    \subfigure[$w_{50}^{k}(t), k \in \{7, \dots, 12\}$ of TFT]{\includegraphics[width=0.48\linewidth]{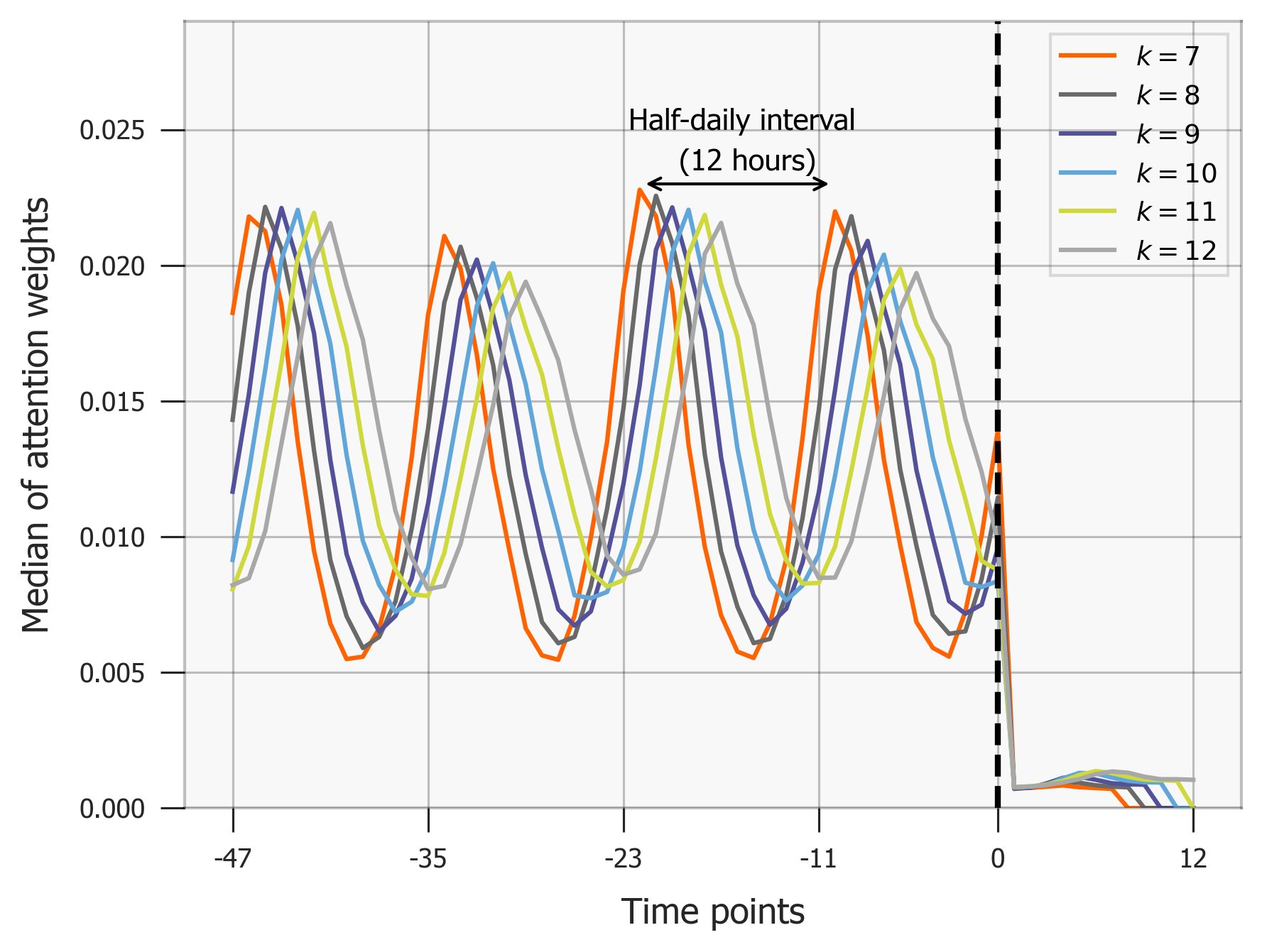}}
    \caption{
    Time trend of impact of variables on forecasting of future time point $k$. The upper row presents the median ($50$th percentiles) of the attention weights obtained from InstaTran. The lower row presents the relevant attention weights from TFT.}
    \label{fig:tp}
\end{figure}
 
Figures \ref{fig:tp} (a) and (b) display the median attention weights $w_{50}^k(t)$ for $k \in \{1, \cdots, 6\}$ and  $k \in \{7, \cdots, 12\}$, respectively.  Correspondingly, Figure \ref{fig:tp} (c) and (d) present the relevant weights obtained from TFT. These figures illustrate time trends of the impact of variables on the forecasting of future time point $k$ for various $k$ values.
For instance, $w_{50}^1(t)$ exhibits the impact of feature variables over time $t \in \{-47, \cdots, 12\}$ on predicting one hour ahead of the input time. This weight, corresponding to InstaTran presented in Figure \ref{fig:tp} (a), demonstrates a 12-hour periodic pattern. Similarly, $w_{50}^2(t)$ in the same figure represents the impact of feature variables over time on the forecasting of two hours ahead and also shows the 12-hour periodic pattern. The patterns observed in $w_{50}^2(t)$ exhibit an approximate one-hour delay in comparison to $w_{50}^1(t)$, revealing a similar periodic pattern akin to that of $w_{50}^1(t-1)$.
The 12-hour periodic trend, as well as the one-hour delay trend, holds for the general case of $w_{50}^{k}(t)$. The same observations apply to the TFT results.
The consistent 12-hour periodic patterns, marked by nearly equidistant time intervals for $k$, indicate that the behavior of the tidal river is captured by both InstaTran and TFT.
Notably, both Instatran and TFT exhibit an evident decrease in weight magnitudes after $t=0$. This reduction is expected, given that these weights correspond to future time points, where the available information for weight construction is inherently limited. In contrast, earlier time points have access to complete information.

While the weight magnitudes of both InstaTran and TFT decrease after $t=0$, InstaTran maintains a substantially larger magnitude in comparison to TFT, particularly for larger $k$ values. Additionally, the 12-hour periodic pattern after $t=0$ is clearer in InstaTran.
This discrepancy highlights InstaTran's strengthened capability to capture and leverage the tidal trend for forecasting future time points, aligning with established scientific evidence. Consequently, this observation enhances the argument that the causal relations embedded within the InstaTran architecture effectively capture the underlying causal phenomena in play.

\subsection{Comparative evaluation of forecasting performance compared to other probabilistic forecasters \label{sec:comp}}

\begin{figure}[!t]
    \centering
    \subfigure[DeepAR]{\includegraphics[width=0.23\linewidth]{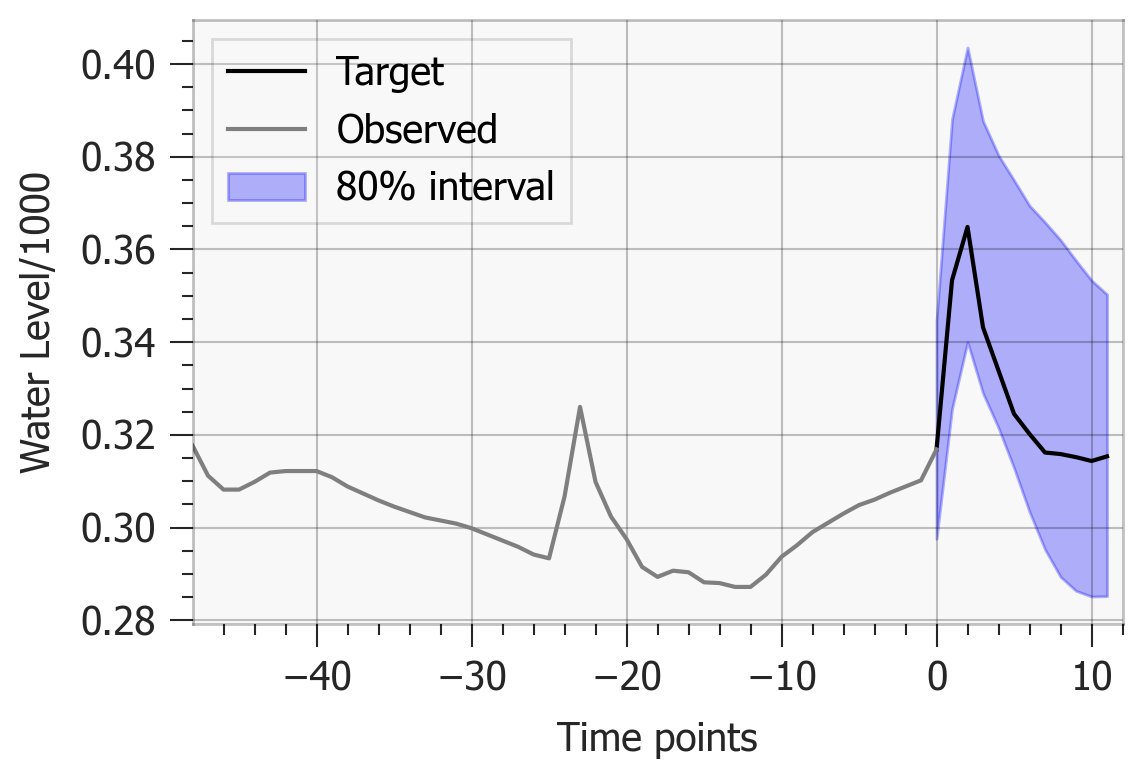}}
    \subfigure[MQ-RNN]{\includegraphics[width=0.23\linewidth]{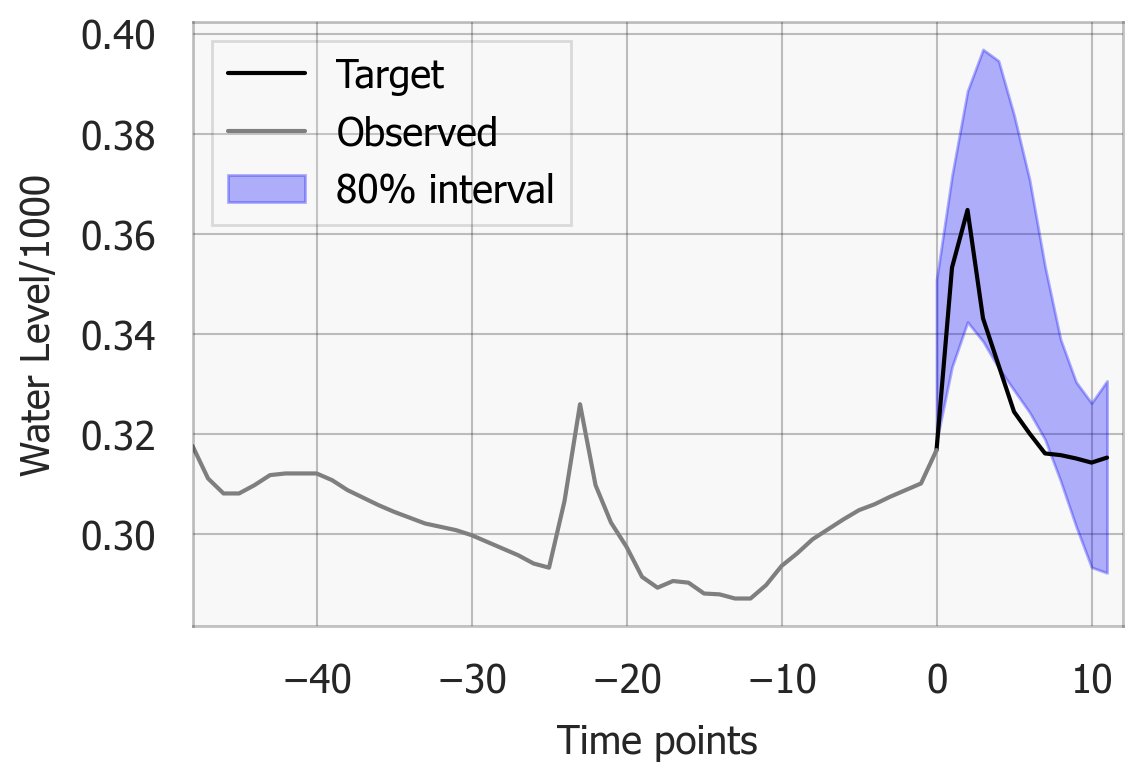}}
    \subfigure[TFT]{\includegraphics[width=0.23\linewidth]{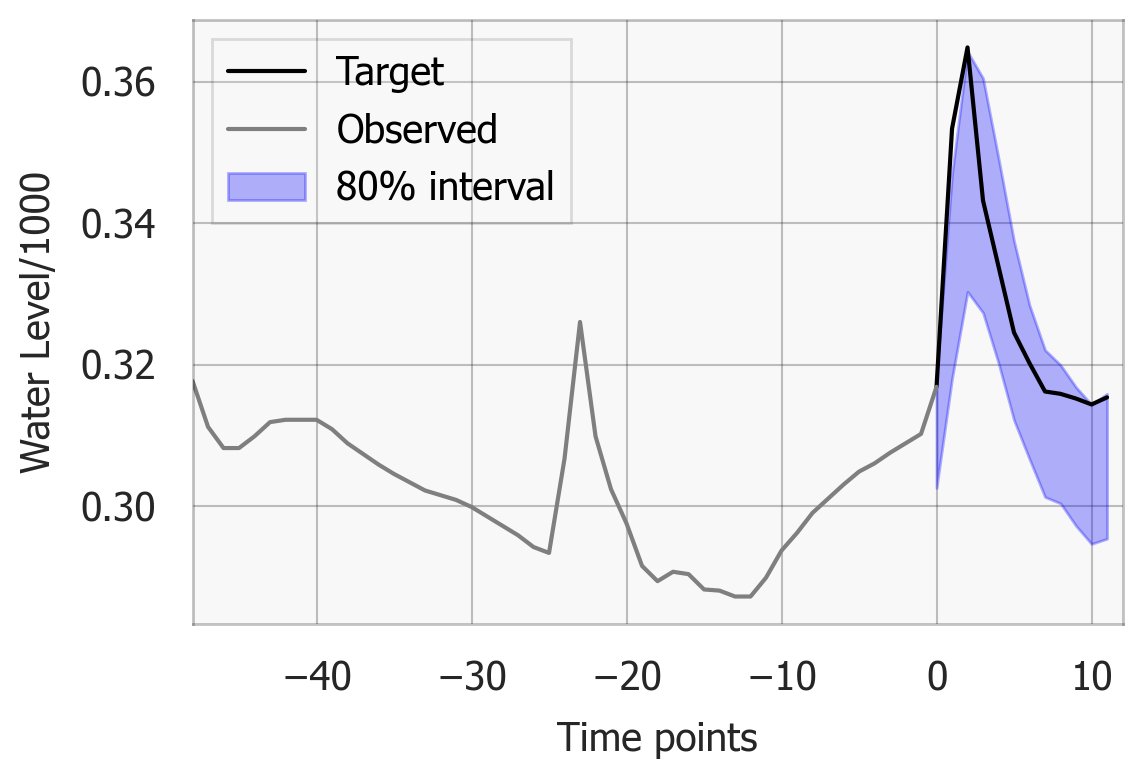}}
    \subfigure[InstaTran]{\includegraphics[width=0.23\linewidth]{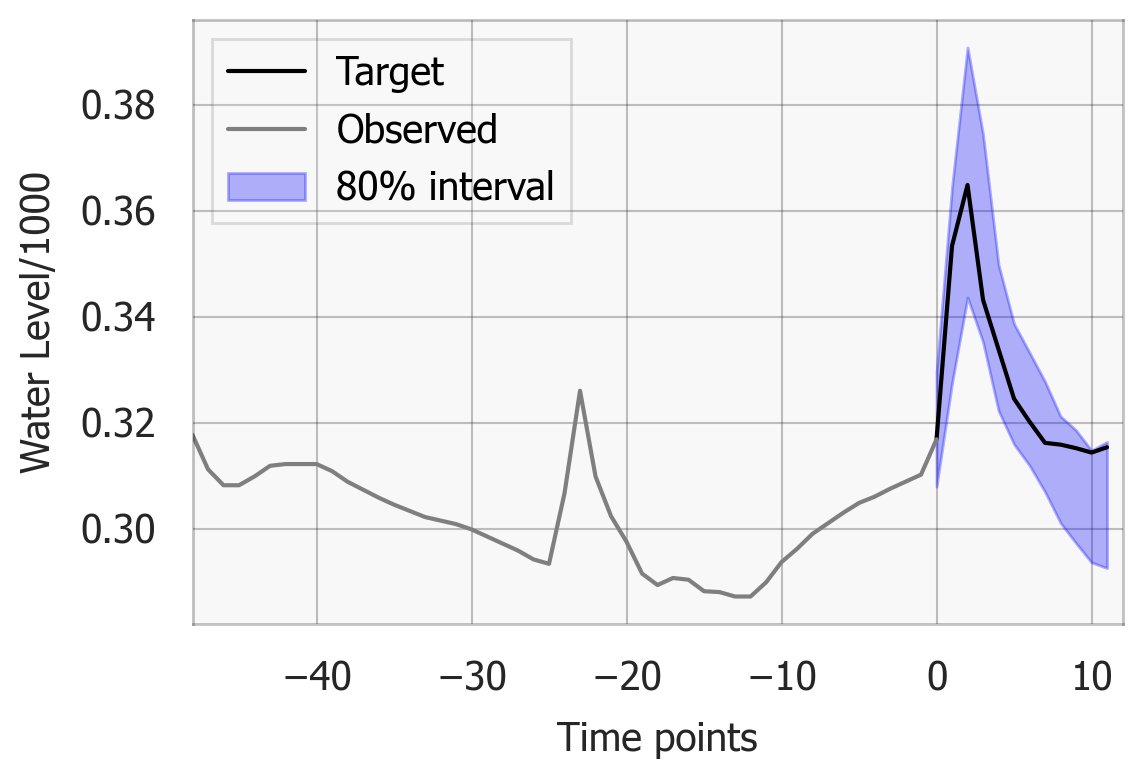}}
    \caption{An example of forecasting results at a selected time point in the test set.}
    \label{fig: results_ex}
\end{figure}

\begin{table}[t]
    \centering
    \caption{Performance comparison of InstaTran and benchmark methods.  The most favorable outcomes are indicated in bold. The second and third ranks are underlined and double-underlined, respectively.}
    \begin{adjustbox}{width=\textwidth}
    
    \begin{tabular}{c c c c c c c} 
    \Xhline{1.5pt}  
    Metric & $q$ & ETS & ARIMA & Theta & LightGBM & STA-LSTM \\ \hline 
    \multirow{3}{*}{average $q$\mbox{-level QL}} & 0.9 & 0.0071 & 0.0110 & 0.0078 & \textbf{0.0016} & 0.0028 \\
    & 0.7 & 0.0118 & 0.0163 & 0.0138 & \textbf{0.0025} & 0.0049 \\
    & 0.5 & 0.0141 & 0.0182 & 0.0138 & \textbf{0.0028} & 0.0052 \\ \hline
    \multirow{3}{*}{$q$-Rate ($|q - q\mbox{-}\mbox{Rate}|$)} & 0.9 & 0.851 (0.049) & 0.652 (0.248) & 0.943 (0.043) & \textbf{0.905 (0.005)} & \textbf{0.905 (0.005)} \\
    & 0.7 & 0.797 (0.097) & 0.307 (0.393) & 0.863 (0.163) & \dunderline{0.749 (0.049)} & \underline{0.728 (0.028)} \\
    & 0.5 & 0.751 (0.251) & 0.057 (0.443) & 0.740 (0.240) & \dunderline{0.622 (0.122)} & \textbf{0.547 (0.047)} \\
    \Xhline{1.5pt}   
    \end{tabular}
    
    \end{adjustbox}
    
    \bigskip
    \begin{adjustbox}{width=\textwidth}
    \begin{tabular}{c c c c c c c} 
    \Xhline{1.5pt}  
    Metric & $q$ & HSDSTM & DeepAR & MQ-RNN & TFT & InstaTran \\ \hline 
    \multirow{3}{*}{average $q$\mbox{-level QL}} & 0.9 & 0.0023 & 0.0027 & 0.0030 & \underline{0.0019} & \dunderline{0.0021}  \\
    & 0.7 & 0.0040 & 0.0044 & 0.0051 & \underline{0.0031} & \dunderline{0.0036} \\
    & 0.5 & 0.0043 & \underline{0.0039} & 0.0053 & \underline{0.0033} & \dunderline{0.0040} \\ \hline
    \multirow{3}{*}{$q$-Rate ($|q - q\mbox{-}\mbox{Rate}|$)} & 0.9 & 0.941 (0.041) & 0.970 (0.070) & 0.930 (0.030) & 0.870 (0.030) & \dunderline{0.924 (0.024)} \\
    & 0.7 & 0.808 (0.108) & 0.925 (0.225)& 0.788 (0.088) & \textbf{0.708 (0.008)} & 0.796 (0.096) \\
    & 0.5 & 0.630 (0.130) & 0.788 (0.288) & 0.625 (0.125) & \underline{0.392 (0.108)} & 0.647 (0.147) \\
    \Xhline{1.5pt}   
    \end{tabular}
    \end{adjustbox}
    \label{tab: results}
\end{table}

We assess both the benchmark models and our proposed InstaTran using two metrics, as defined in  \eqref{def: ql_loss} and \eqref{def: qrate}. The performance results on the test dataset are presented in Table \ref{tab: results}, demonstrating that LightGBM and TFT outperform other models in most aspects. However, the proposed InstaTran yields competitive results with the third rank in QL measures among the ten models evaluated.

Figure \ref{fig: results_ex} presents visualizations of forecasted outcomes from the test set for a specific $\bu$ selection. The past observations and the target water level of $B_0$ are indicated by the gray and black lines, respectively. The blue band encompasses predicted quantiles ranging from $0.1$ to $0.9$.
The intervals generated by DeepAR are notably wider, which diminishes their precision and may result in potentially inconclusive findings.
The MQ-RNN generates an interval that deviates from the intended target levels.
In contrast, TFT and InstaTran provide narrower and more dependable intervals, with InstaTran particularly demonstrating reliability around the sharp peak of the target.

For a more extensive evaluation, we performed a similar data analysis for water level prediction on the US lake dataset, which includes three lakes--Mead, Mohave, and Havasu--from 2005 to 2022. The proposed method demonstrated competitive results, aligning with those observed in the Han River analysis. Details are provided in the Appendix.

\subsection{Evalutaion of robustness to distribution shift}
This section further explores the performances of InstaTran under distribution (or covariate) shift scenarios. 
Specifically, we investigate distributional shift scenario by splitting the year-round datasets into two parts: the observations from May and June between 2016 and 2021 are utilized as the training set, and those from July and August from the corresponding years as the test set, in which fitting is performed in a year-specific manner. In the area, rainfall has a seasonal pattern that differs between May-June and July-August. May and June typically exhibit mild precipitation patterns, while July and August feature notably higher precipitation (see the Appendix).

\begin{table}[t]
    \centering
    \caption{Performance comparison of InstaTran and benchmark methods.  The most favorable outcomes are indicated in bold. The second rank is underlined. The second and third ranks are underlined and double-underlined, respectively.}
    \begin{adjustbox}{width=\textwidth}
    \begin{tabular}{c c c c c c c} 
    \Xhline{1.5pt}  
    Metric & $q$ & ETS & ARIMA & Theta & LightGBM & STA-LSTM \\ \hline 
    \multirow{3}{*}{average $q$\mbox{-level QL}} & 0.9 & 0.0160 & 0.0418 & \underline{0.0093} & 0.0254 & 0.0193 \\
    & 0.7 & 0.0175 & 0.0411 & 0.0148 & 0.0240 & 0.0392 \\
    & 0.5 & 0.0162 & 0.0345 & 0.0159 & 0.0188 & 0.0520 \\ \hline
    \multirow{3}{*}{$q$-Rate ($|q - q\mbox{-}\mbox{Rate}|$)} & 0.9 & 0.501 (0.376) & 0.523 (0.434) & 0.773 (0.159) & 0.628 (0.272) &\underline{0.894 (0.093)} \\
    & 0.7 & 0.405 (0.326) & 0.421 (0.384) & 0.553 (0.255) & 0.471 (0.229) & 0.782 (0.187) \\
    & 0.5 & 0.343 (0.300) & 0.338 (0.318) & 0.351 (0.298) & \dunderline{0.385 (0.130)} & 0.681 (0.284) \\
    \Xhline{1.5pt}   
    \end{tabular}
    \end{adjustbox}
    
    \bigskip
    \begin{adjustbox}{width=\textwidth}
    \begin{tabular}{ c c c c c c c} 
    \Xhline{1.5pt}  
    Metric & $q$ & HSDSTM & DeepAR & MQ-RNN & TFT & InstaTran \\ \hline 
    \multirow{3}{*}{average $q$\mbox{-level QL}} & 0.9 & 0.0315 & 0.0097 & \dunderline{0.0094} & 0.0109 & \textbf{0.0087} \\
    & 0.7 & 0.0818 & \textbf{0.0109} & \dunderline{0.0154} & 0.0160 & \underline{0.0117} \\
    & 0.5 & 0.1433 & 0.0121 & \underline{0.0120} & 0.0199 & \textbf{0.0119} \\ \hline
    \multirow{3}{*}{$q$-Rate ($|q - q\mbox{-}\mbox{Rate}|$)} & 0.9  & \dunderline{0.851 (0.094)} & 0.608 (0.271) & \textbf{0.950 (0.070)} & 0.697 (0.209) & 0.725 (0.170) \\
    & 0.7 & 0.774 (0.193) & \textbf{0.607 (0.111)} & 0.594 (0.199) & \dunderline{0.610 (0.176)} & \underline{0.622 (0.127)} \\
    & 0.5 & 0.714 (0.287) & \underline{0.606 (0.128)} & 0.520 (0.152) & 0.565 (0.189) & \textbf{0.540 (0.107)} \\
    \Xhline{1.5pt}   
    \end{tabular}
    \end{adjustbox}
    \label{tab: ds_results}
\end{table}

\begin{figure}[!t]
    \centering
    \subfigure[Variable importance of $P_1$.]{\includegraphics[width=0.48\linewidth, height=4.5cm]{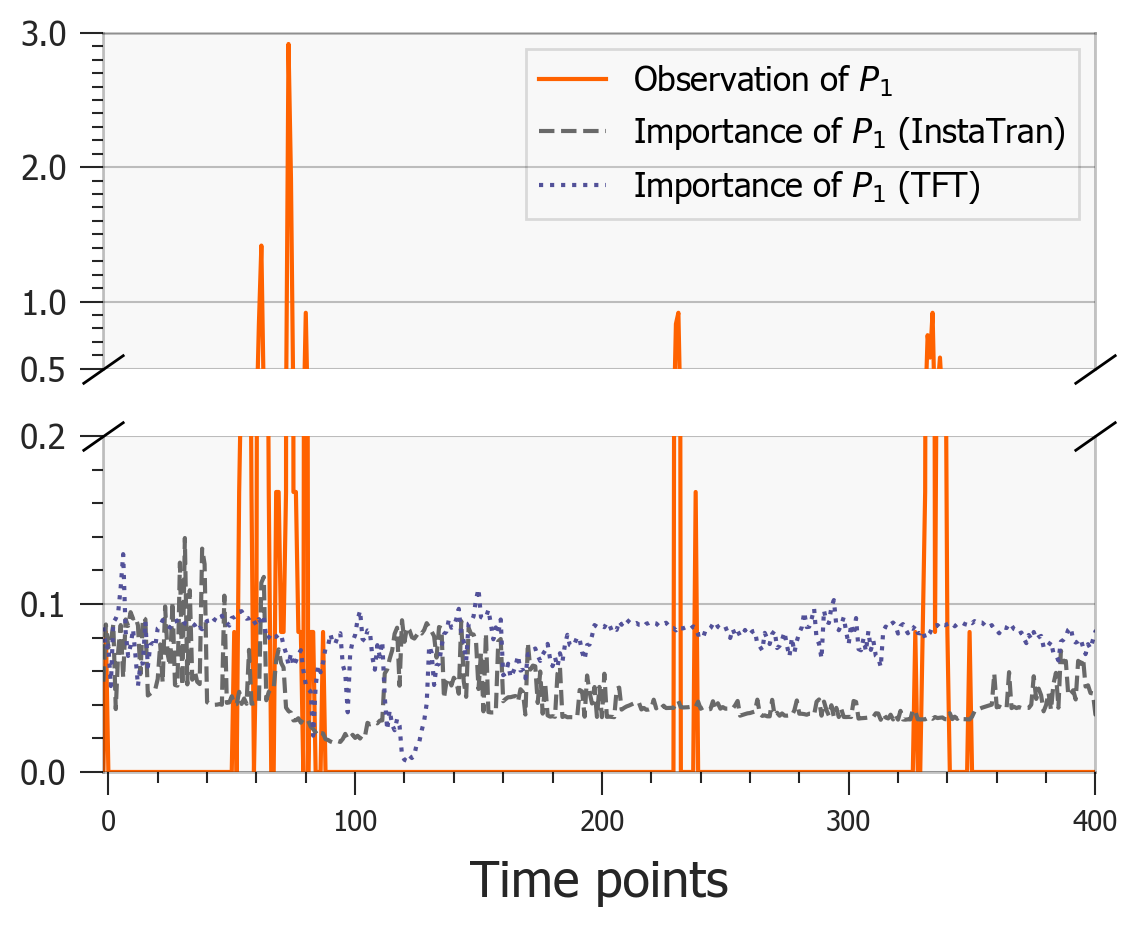}}
    \subfigure[Water level forecasting results.]{\includegraphics[width=0.48\linewidth]{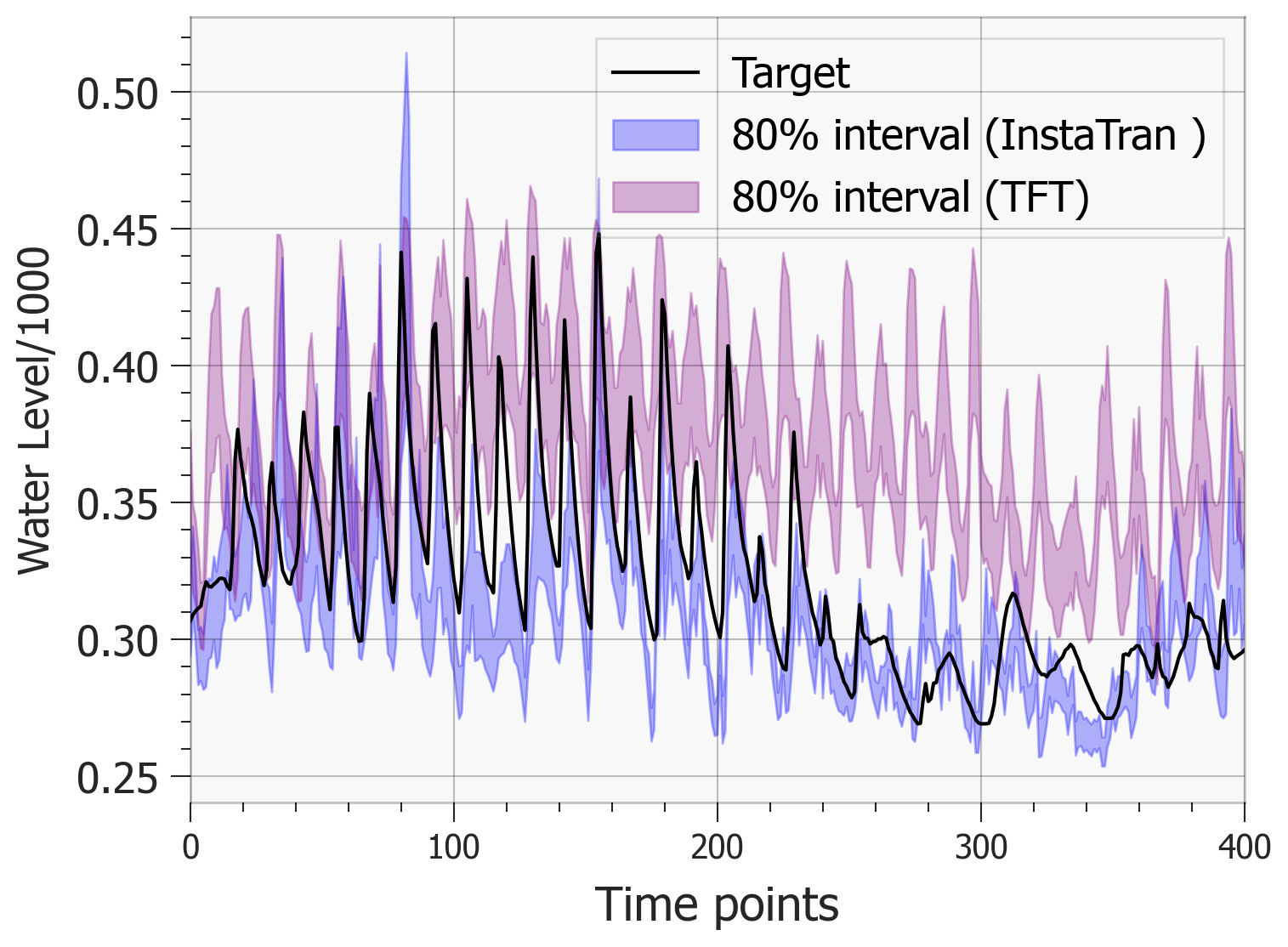}}
    \caption{Variable importances and observations of $P_1$ (left), and forecasting results of InstaTran and TFT (right) under the distribution shift scenario.}
    \label{fig: dist_shift}
\end{figure}

Table \ref{tab: ds_results} presents the average performance in July and August across the years 2016 to 2021. While LightGBM and TFT demonstrate superior performance among benchmarks under stable distribution scenario in Section \ref{sec:comp}, it is notable that its performance degrades, even trailing behind that of Theta.
On the other hand, the proposed InstaTran exhibits strong performance under this scenario across all metrics.
Also, simpler deep learning models exhibit robustness to distribution shift and surpass TFT in performance. Conversely, domain-specific models such as STA-LSTM and HSDSTM are susceptible to the distribution shift.

This observation suggests that TFT, LightGBM, and domain-specific models may overly adapt to the training data, resulting in superior performance under stable year-round rainfall patterns in Section \ref{sec:comp}, but faltering when faced with differing conditions as presented in this section.
In Section \ref{sec:comp}, observations from 2016 to 2020 were used as the training set and those from 2021 as the test set. The close similarity in distribution between the training and test sets due to the stable year-round rainfall pattern makes this setting more favorable for methods that fully adapt to training data.
The proposed method, InstaTran, demonstrates robust performance and mitigates overfitting
which indicates that it has successful encoding of the predefined causal relationship into the prediction process.

In the preceding Section 4.2, the interpretations of predictors for TFT and LightGBM indicate that these models may prioritize historical records and follow past patterns for future predictions rather than actively reflecting causal relationships. This observation is further supported by comparing the importance of $P_1$ and $\text{WL}(B_0)$ in InstaTran and TFT.\footnote{While the feature importance of these variables from LightGBM is available, its metric differs from that of variable importance, and thus we do not include a comparison between LightGBM and InstaTran.}
Figures \ref{fig: dist_shift} (a) and (b) display the variable importance of $P_1$ and the forecasting results of $\mbox{WL }(B_0)$, respectively. In Figure \ref{fig: dist_shift} (a), the importance of $P_1$ is overestimated over mild rain season (May and June) in the case of TFT compared to InstaTran. 
This overestimation of the effects of $P_1$ leads to an overestimation of water level in the rainy season (July and August), as observed in Figure \ref{fig: dist_shift} (b).

In natural science, distribution shifts are frequently encountered \citep{chadwick2022regional}, rendering forecasters trained solely on a particular distribution ineffective \citep{fan2023dish}. 
To address this challenge, prior studies have demonstrated that coefficients grounded in causality can yield robust results even in the face of distribution shifts \citep{rojas2018invariant, rothenhausler2021anchor}. Additionally, \cite{mitrovic2021representation} argued that representations based on a causal framework can enhance generalization capabilities in scenarios involving distribution shifts. 
By incorporating a predefined causal structure, 
our proposed representation learning approach enhances the robustness of the forecaster to distribution shift and accurately estimates the importance of variables.

\section{Conclusion and Limitation}
We proposed a deep learning architecture for multiple quantile forecasting with spatiotemporal causal structure. Our proposed architecture extended the capabilities of the existing transformer by incorporating spatial and temporal masks that encode causal relations. This approach enabled the model to incorporate prior knowledge into its feature learning process, yielding results that are in alignment with established understandings. 
Moreover, it provided a convenient mechanism to evaluate the effective integration of inputted causal relation knowledge by examining the resulting attention layer weights. The magnitudes of these weights served as a measure of importance, providing a practical means to assess the importance of each variable.
In the decoding step, our proposed approach simultaneously predicted multiple quantiles for various time points, mitigating the risk of error accumulation.

We conducted water level forecasting studies for the Han River employing our proposed model and analyzed the resulting attention weights with a focus on their interpretability. 
The resulting variable importance of the proposed model aligned with the presumed causal relations in both spatial and temporal domains, effectively embedding the causality built upon physical laws and established understandings present in the input.
Furthermore, the temporal attention weights of the proposed method effectively captured inherent periodic patterns within the nature of the response variable, while the periodic pattern itself was not explicitly modeled. 
In addition, our proposed method not only yielded highly interpretable results that align with existing understanding but also enhanced the robustness of the forecaster to distribution shift scenarios. A supplementary analysis of the US lake data in the Appendix further supports that our proposed model remains a strong benchmark, demonstrating the robustness and generalizability of our approach.

The proposed approach substantially enhances the applicability of deep learning models by offering avenues for integrating prior causal relation knowledge and facilitating the interpretation of how effectively such input knowledge is captured.
However, certain limitations pertain to the types of input causal relations that can be integrated.
The proposed model is designed to accommodate relatively straightforward causality structures, such as directional graphs representing variable dependencies or irreversibility for temporal dependencies.
The general causal structure in latent space cannot be embedded in the current approach.
It is expected that research in the nonlinear structural causal model would enable the construction of a more comprehensive, interpretable deep learning model. This aspect of the research is left for future investigations.

\section*{Author Credit Statement}
\textbf{Sungchul Hong}: Conceptualization, Methodology, Software, Data Curation, Visualization, Writing - Original Draft. \textbf{Yunjin Choi}: Validation, Investigation, Writing - Original Draft, Writing - Review \& Editing. \textbf{Jong-June Jeon}: Writing - Original Draft, Supervision, Project administration, Funding acquisition.

\section*{Acknowledgements}
All authors were supported by the National Research Foundation of Korea grant [NRF-2022R1A4A3033874]. 
Sungchul Hong was supported by the National Research Foundation of Korea grant [NRF-2022M3J6A1084845]
Jong-June Jeon was supported by the National Research Foundation of Korea grant [NRF-2022R1F1A1074758].
The authors acknowledge the Urban Big data and AI Institute of the University of Seoul supercomputing resources (http://ubai.uos.ac.kr) made available for conducting the research reported in this paper. Additionally, the authors extend their gratitude to dacon.io for providing the dataset used in this research.

\section*{Declaration of Interests}
The authors declare that they have no known competing financial interests or personal relationships that could have appeared to influence the work reported in this paper.

\bibliography{wlf}

\clearpage

\appendix

\section{Descriptive statistics of variables.}
Table \ref{tab:dec_stat} presents the basic statistics of variables. The observations of precipitations have most $0$, but their maximum values are too big to have a high impact. There are some variables that have a larger standard deviation than the mean, such as IF$(D)$ and variables related to flow. 
\begin{table}[!ht]
    \centering
    \caption{Descriptive statistics of variables in the Han River dataset. *: The minimum value of FL($B_3$) can be negative due to barrages in the Han River or tide.}
    \begin{adjustbox}{width=\textwidth}
    \begin{tabular}{c c c c c c c c} 
    \Xhline{1.5pt}  
        & Mean & Std & Min & 25\% & Median & 75\% & Max\\ \hline 
    \text{$P_1$} & 0.03 & 0.24 & 0.0 & 0.0  & 0.0  & 0.0 & 8.5 \\
    \text{$P_2$} & 0.04 & 0.27 & 0.0 & 0.0  & 0.0  & 0.0 & 9.33 \\
    \text{$P_3$} & 0.03 & 0.24 & 0.0 & 0.0  & 0.0  & 0.0 & 8.0 \\
    \text{WL($B_4$)} & 346.49 & 171.89 & 55.33 & 194.83 & 324.50 & 485.33  & 811.33 \\
    \text{WL($D$)}  & 25.04 & 0.14  & 24.13 & 24.94 & 25.04 & 25.14 & 25.42 \\
    \text{IF($D$)}  & 590.45 & 1213.37 & 0.0 & 136.0 & 269.61 & 510.39   & 18830.0 \\
    \text{STR($D$)} & 212.71 & 5.32 & 178.37 & 209.36 & 212.92 & 216.46 & 226.46 \\
    \text{JUS($D$)} & 31.29 & 5.32 & 17.55 & 27.54 & 31.08 & 34.64 & 65.64 \\
    \text{OF($D$)}  & 582.60 & 1213.05 & 0.0 & 134.0 & 216.17 & 503.0 & 18161.67 \\
    \text{WL($B_0$)} & 332.49 & 82.28 & 260.7 & 288.7 & 309.2 & 346.7 & 1287.2 \\
    \text{WL($B_1$)} & 319.79 & 72.96 & 250.37 & 279.87 & 300.2 & 334.37 & 1142.87 \\
    \text{FL($B_1$)} & 784.05 & 1152.50 & 243.29 & 325.73 & 458.62 & 764.20 & 9405.6 \\
    \text{WL($B_2$)} & 317.02 & 68.13 & 252.0 & 278.0 & 299.0 & 332.17 & 1067.67 \\
    \text{FL($B_2$)} & 640.25 & 1503.29 & -3118.98* & 215.05 & 373.21 & 720.21 & 24859.13 \\
    \text{WL($B_3$)} & 303.14 & 54.68 & 242.3 & 269.13 & 289.05 & 320.47 & 839.47 \\
    \text{FL($B_3$)} & 1130.82 & 1623.81 & 219.52 & 469.24 & 718.04 & 1219.26 & 29501.66 \\
    \Xhline{1.5pt}   
    \end{tabular}
    \end{adjustbox}
    \label{tab:dec_stat}
\end{table}

\section{Hyperparameter setting}
\begin{table}[!ht]
    \centering
    \caption{Hyperparameter we considered of InstaTran and deep learning-based benchmark models.}
    \begin{adjustbox}{width=0.95\textwidth}
    \begin{tabular}{l c c c c c c} 
    \Xhline{1.5pt}  
    Model & Hidden unit dimension & Embedding dimension & LSTM layers &  Dropout ratio & Batch size & Epochs \\ \hline 
    STA-LSTM  & $48$ & -   & $1$ & $0.1$ & $500$ & $100$ \\
    HSDSTM    & $16$ & -   & $2$ (TCN layers) & $0.1$ & $500$ & $100$ \\    
    DeepAR    & $30$ & $3$ & $3$ & $0.1$ & $500$ & $200$ \\
    MQ-RNN    & $5$  & $1$ & $3$ & $0.1$ & $500$ & $200$ \\
    TFT       & $30$ & $5$ & $1$ & $0.1$ & $500$ & $150$ \\
    InstaTran & $10$ & $3$ & -   & $0.1$ & $500$ & $200$ \\
    \Xhline{1.5pt}   
    \end{tabular}
    \end{adjustbox}
    \label{tab:hyperparam}
\end{table}

\clearpage
\section{Variable Selection Network (VSN)}
\cite{lim2021temporal} constructs the context vector by the VSN that averages over hidden states with trainable weights. Let $w(\phi_{\bu'}) \in \mathbb{R}^{p}$ be trainable weight vector of $\phi_{\bu'} \in \mathbb{R}^{d_0}$. Then, for $\bu'=\bu-B+1, \dots, \bu$, the output of the VSN layer is given by
\bean
\mbox{VSN}(\phi_{\bu'}) = \mathbf{1}_p \mbox{diag}(w(\phi_{\bu'}))\phi_{\bu'} \in \mathbb{R}^{d_0}, 
\eean
where $\mathbf{1}_p$ is the row vector whose all elements are ones. 
For details on $w(\phi_{\bu'})$, see Section 4.2 of \cite{lim2021temporal}. The weight vector $w(\phi_{\bu'})$ assumes a role in determining the significance of spatial variables learned from the initial SCAN layer, capable of capturing the significant event within a specific variable. Thus, the VSN  reduces the spatial feature matrix $\phi_{\bu'}$ into the feature vector called a context vector. Let a collection of the context vectors from $\bu-B+1$ to $\bu$ be  
\bean
\Phi_{\bu,B} = [\mbox{VSN}(\phi_{\bu-B+1})^\top , \dots ,  \mbox{VSN}(\phi_{\bu})^\top ]^\top \in \mathbb{R}^{B \times d_0}. 
\eean

\section{Comparing the spatial dependencies with attention weights}
\begin{figure}[t!]
    \centering
    \subfigure[Dry day (STA-LSTM)]{\includegraphics[width=0.93\linewidth]{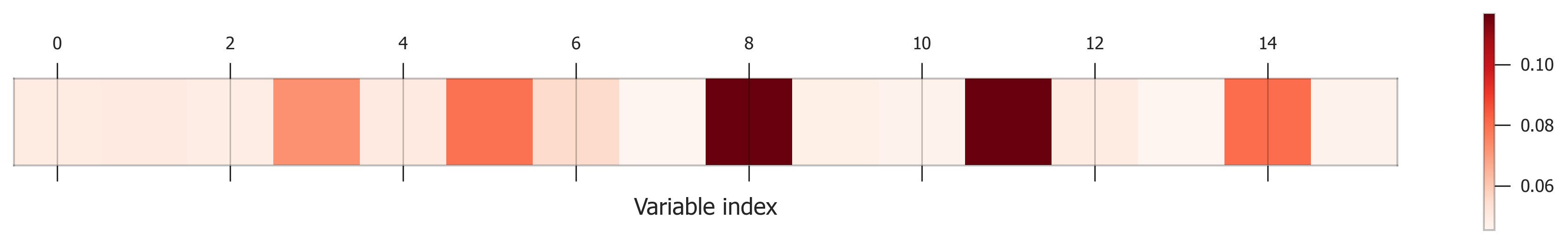}}
    \subfigure[Rainy day (STA-LSTM)]{\includegraphics[width=0.93\linewidth]{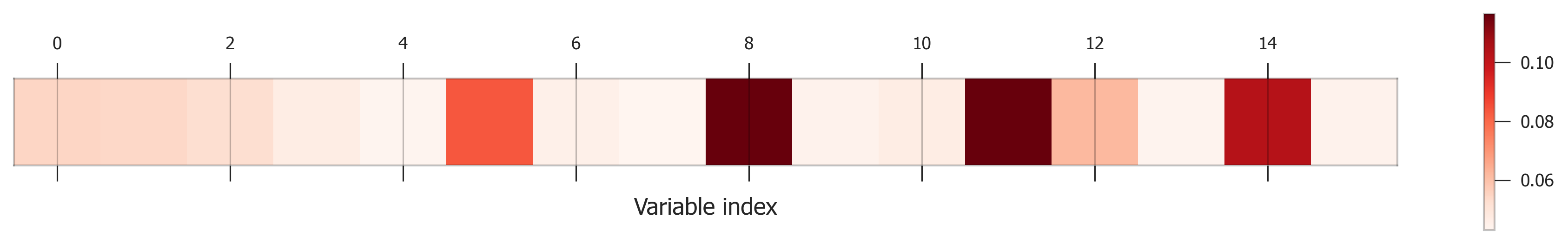}}
    \subfigure[Dry day (HSDSTM)]{\includegraphics[width=0.46\linewidth, height=5cm]{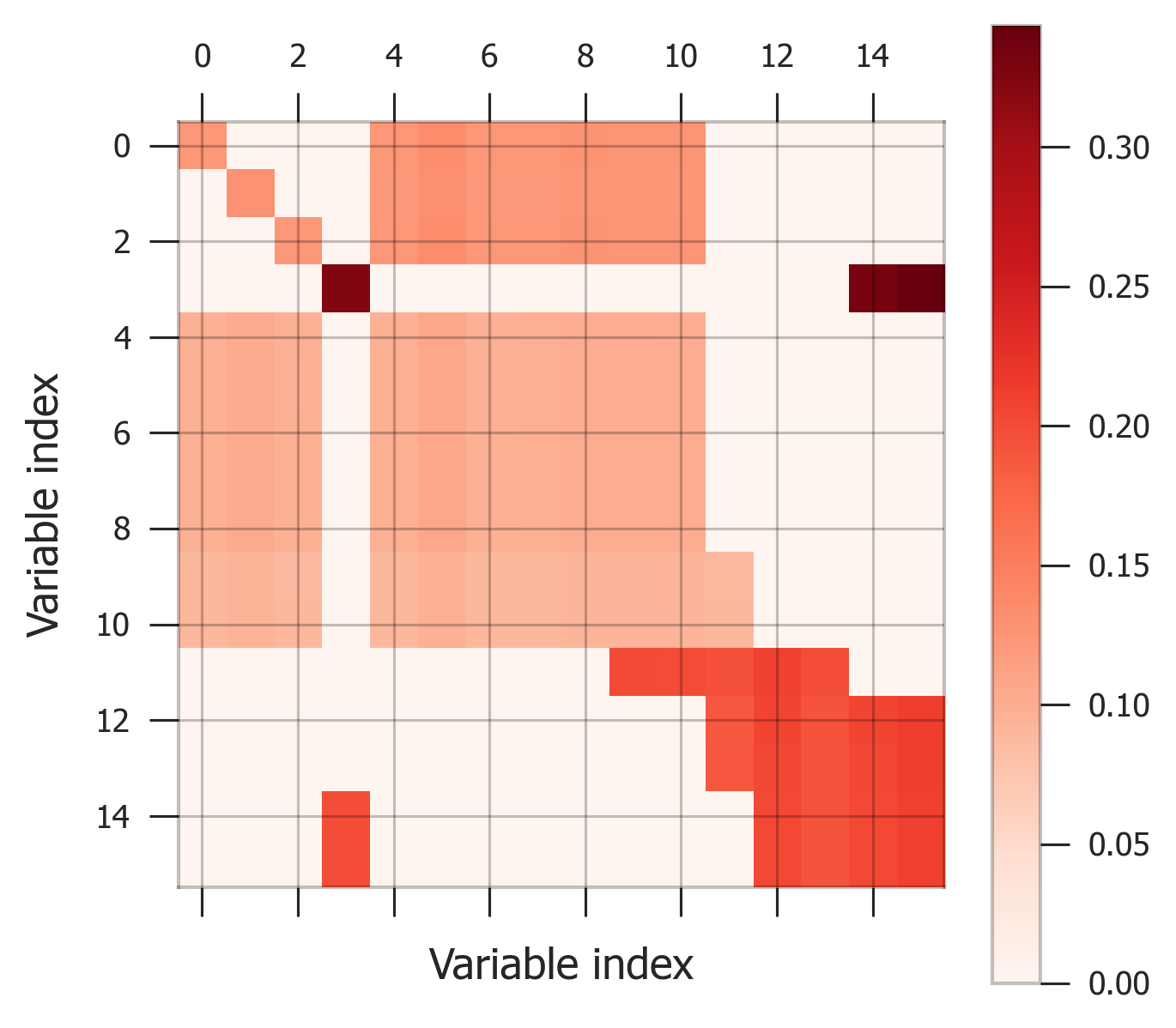}}
    \subfigure[Rainy day (HSDSTM)]{\includegraphics[width=0.46\linewidth, height=5cm]{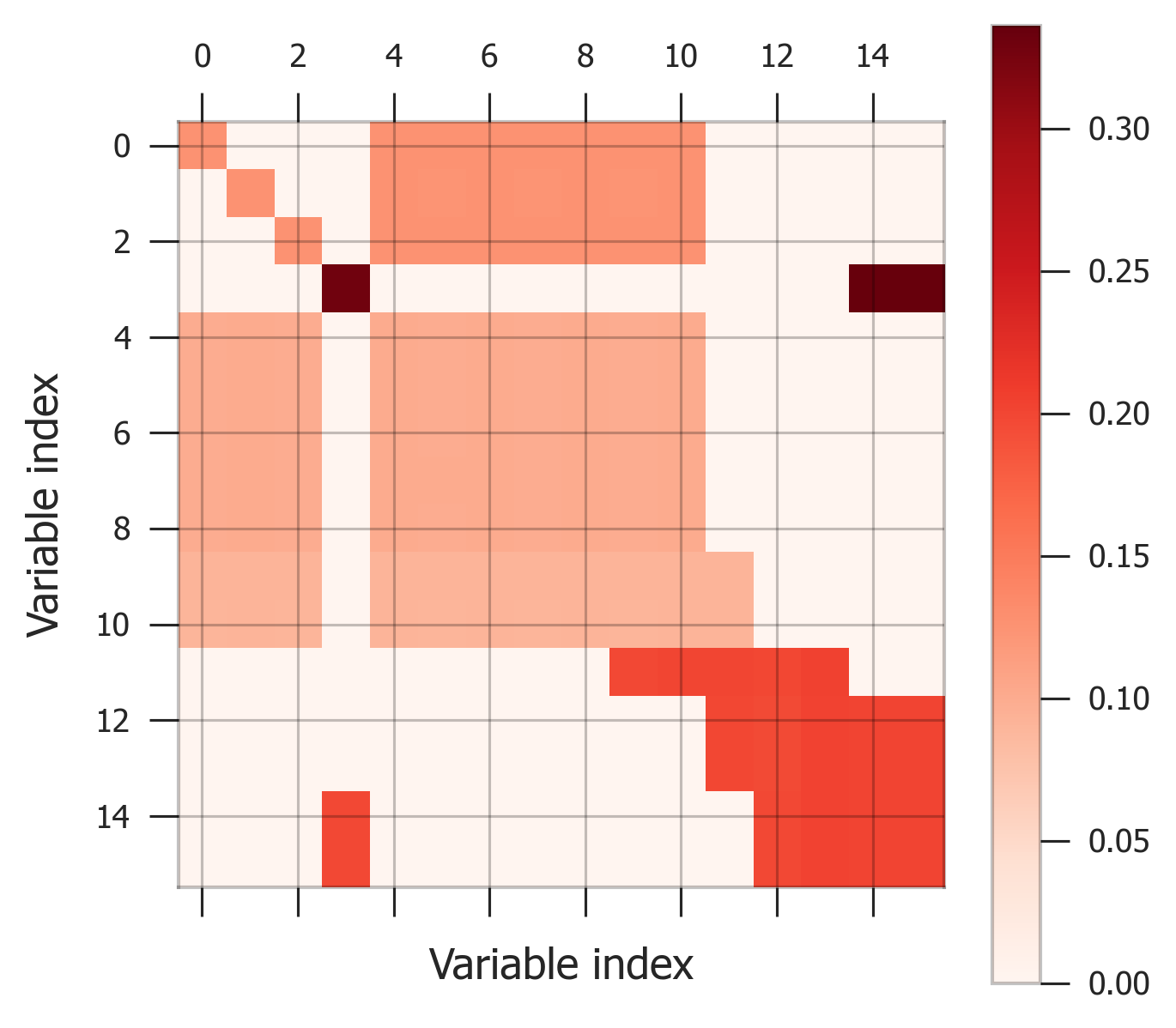}}
    \caption{
    Heatmaps of attention weights and weighted adjacency matrices. Plots (a) and (b) correspond to STA-LSTM, and plots (c) and (d) correspond to HSDSTM. The variables corresponding to the indices of the axes as follows: $0:P_1,~1:P_2, ~2:P_3,~3:\mbox{WL}~(B_4),~4:\mbox{WL}~(D),~5:\mbox{IF}~(D),~6:\mbox{STR}~(D),~7:\mbox{JUS}~(D),~8:\mbox{OF}~(D),~9:\mbox{WL}~(B_1),~10:\mbox{FL}~(B_1),~11:\mbox{WL}~(B_0),~12:\mbox{WL}~(B_2),~13:\mbox{FL}~(B_2),~14:\mbox{WL}~(B_3),~15:\mbox{FL}~(B_3)$.}
    \label{fig: sa_bench}
\end{figure}

We provide the heatmaps illustrating attention weights of STA-LSTM and weighted adjacency matrices of HSDSTM under two scenarios of Section \ref{sec: abl} in Figure \ref{fig: sa_bench}. Figures \ref{fig: sa_bench} (a) and (b) represent the attention weights of a spatial attention module in STA-LSTM. Figures \ref{fig: sa_bench} (c) and (d) represent adaptive weighted adjacency matrices of the first graph attention networks (GAT) in HSDSTM. 
STA-LSTM is limited to analyzing time-varying weights of variables and does not account for spatial dependencies among variables. This limitation makes it challenging to capture the complex dependencies across variables.
The importance of dominant factors, WL ($B_0$) and OF ($D$), is overestimated in both scenarios, similar to the observations in TFT.

In the case of HSDSTM, our analysis reveals that it has difficulty in fully detecting the importance of underlying rainfall variables. This limitation arises since HSDTM relies soley on topological information based on physical models, such as flow path distance, rather than causal relationships that could be influenced by other factors. 

\clearpage
\section{Temporal patterns without $M_{\mathcal{S}}$}
\begin{figure}[ht]
    \centering
    \subfigure[$w_{50}^{k}(j), k \in \{1, \dots, 6\}$]{\includegraphics[width=0.48\linewidth]{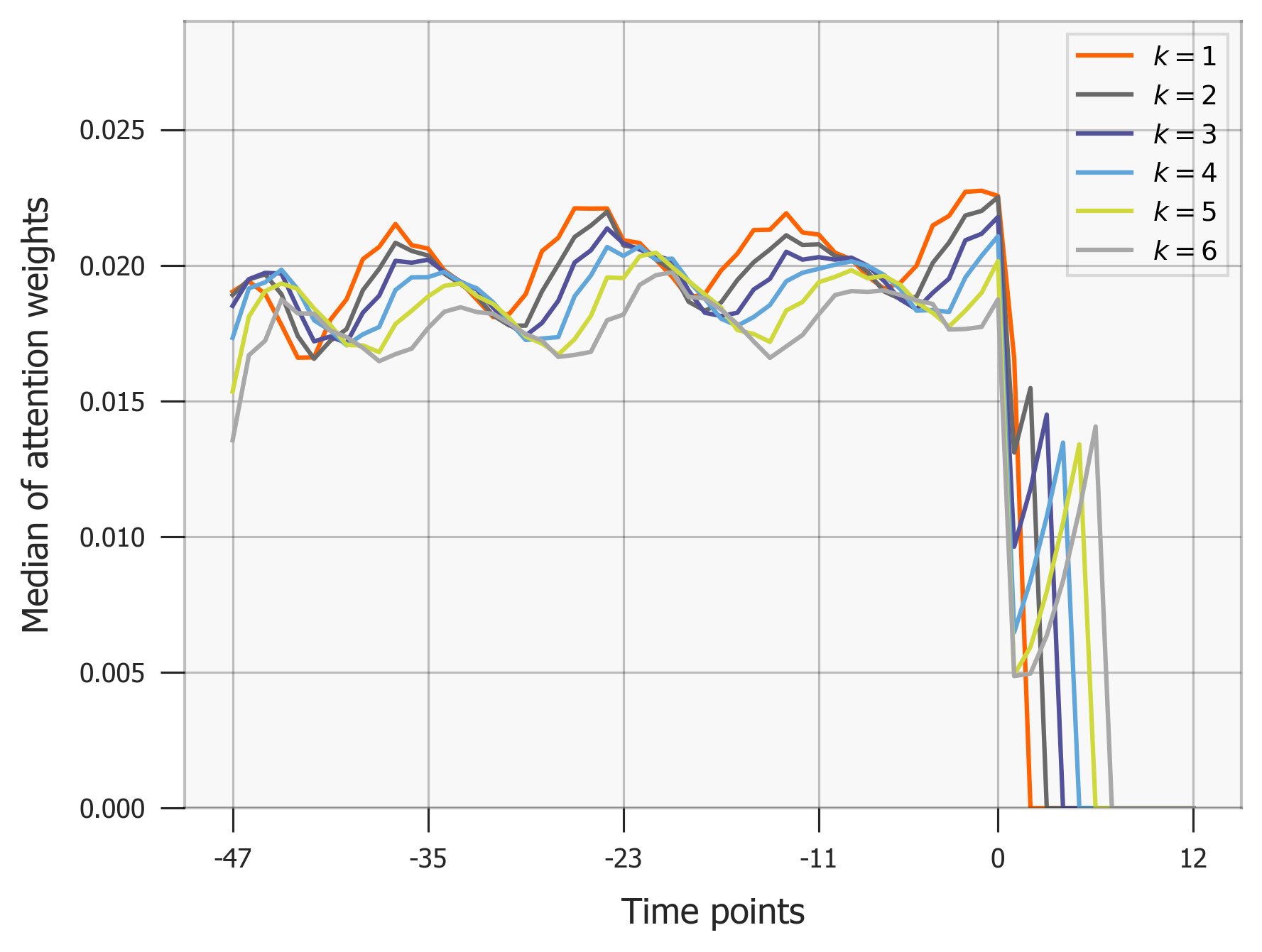}}
    \subfigure[$w_{50}^{k}(j), k \in \{7, \dots, 12\}$]{\includegraphics[width=0.48\linewidth]{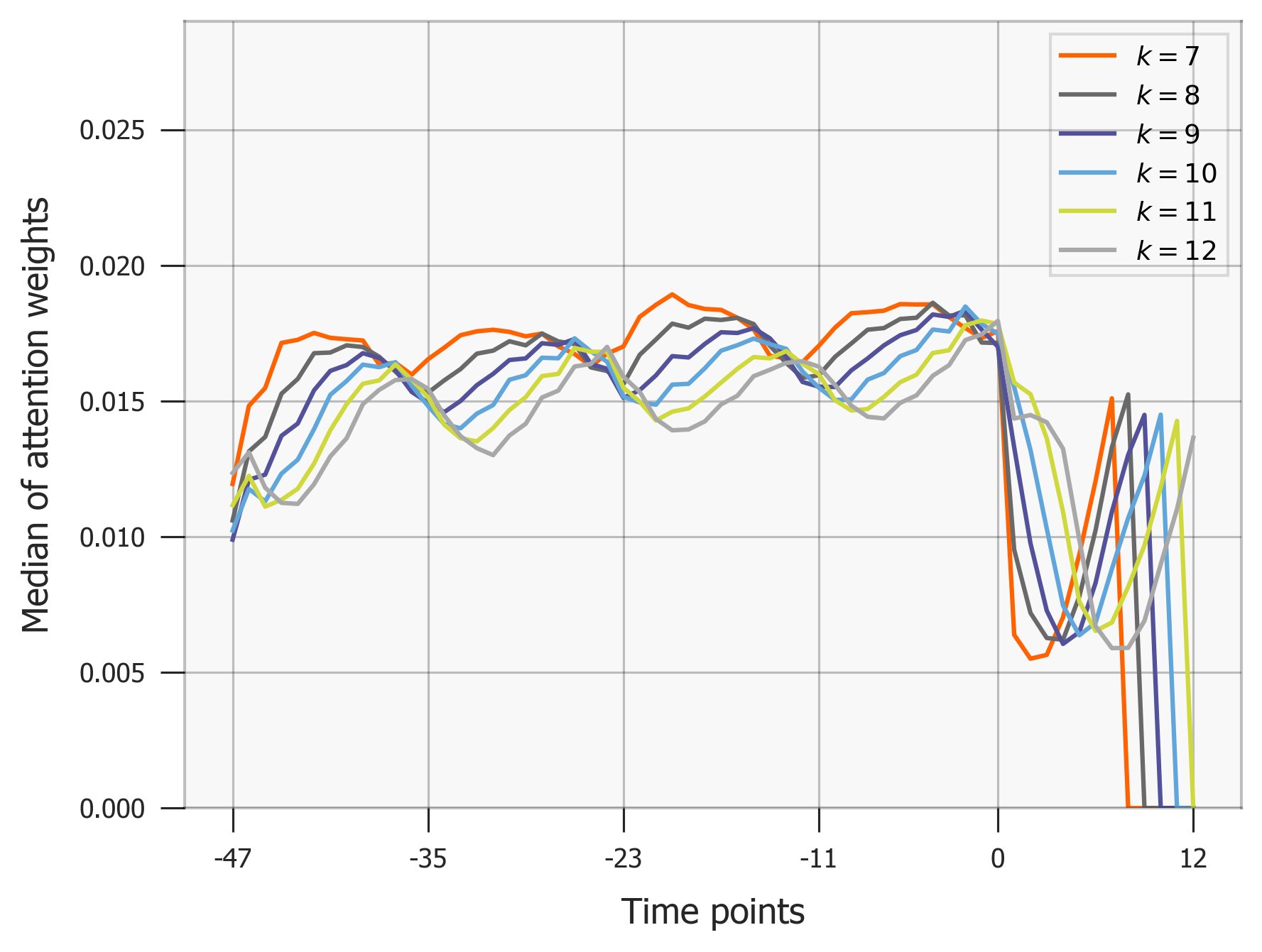}}
    \caption{Time trend of impact of variables on forecasting of future time point $k$ from InstaTran without $M_\mathcal{S}$.}
    \label{fig:tp_without_Ms}
\end{figure}
Figure \ref{fig:tp_without_Ms} shows that the spatial mask $M_\mathcal{S}$ plays an important role in building temporal patterns. In the absence of $M_\mathcal{S}$, the 12-hour periodic pattern becomes unclear, and the significance of future time points becomes similar to past time points, especially in $k \geq 7$. This implies that crucial features are not filtered out from the intricate mixture of variables in the past time points. This inadequate representation learning of past time points leads to poor forecasting performance.

\section{Boxplots of precipitation.}
\begin{figure}[h]
    \centering
    \subfigure[$P_1$]{\includegraphics[width=0.32\linewidth]{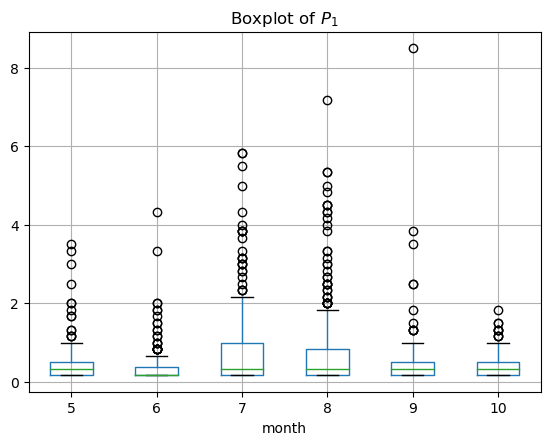}}
    \subfigure[$P_2$]{\includegraphics[width=0.32\linewidth]{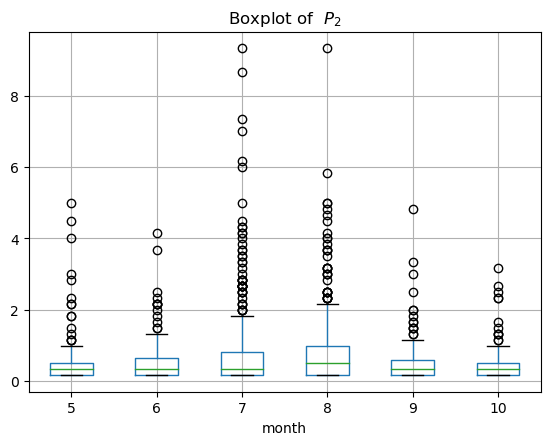}}
    \subfigure[$P_3$]{\includegraphics[width=0.32\linewidth]{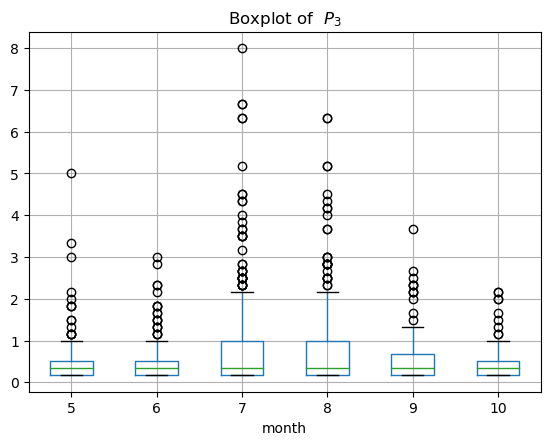}}
    \caption{Boxplots of precipitation variables $P_1$, $P_2$, and $P_3$ with the x-axis representing months.}
    \label{fig:prep}
\end{figure}

\clearpage
\section{Additional experiments with US lakes dataset}
\begin{wrapfigure}{r}{0.44\textwidth}
    \centering
    {\includegraphics[width=0.44\textwidth]{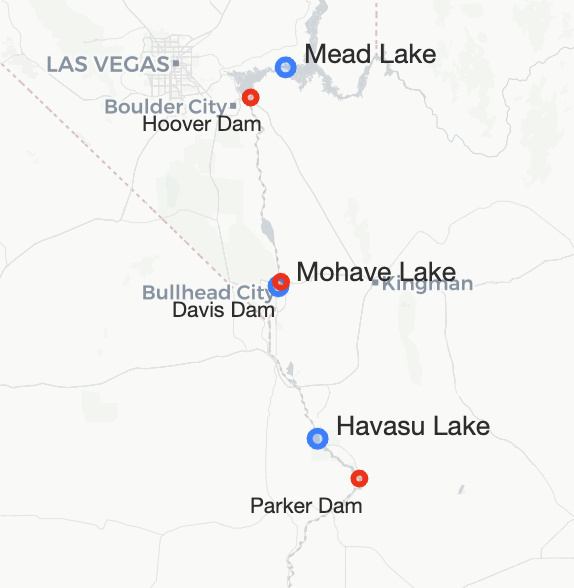}}
    \caption{Map associated with the US lake dataset. The blue and red circles with the labels indicate the locations of lakes and dams, respectively.}
    \label{fig: us_lakes}
\end{wrapfigure}
We extend our analysis to the US lake dataset.\footnote{The US lake dataset is accessible at \url{https://www.water-data.com}.} 
This dataset, spanning from 2005 to 2022, consists of data from three lakes -- Mead, Mohave, and Havasu -- characterized by daily water level, inflow, and outflow. 
The water levels of these lakes are a primary focus, as they are crucial water sources for major drinking water supplies and industries in the US. These reservoirs, associated with the Hoover Dam, are experiencing significant depletion.
The sequence from upstream to downstream is Mead Lake $\rightarrow$ Mohave Lake $\rightarrow$ Havasu Lake. In addition to the water levels, we collected precipitation observations near Mohave and Havasu lakes.\footnote{The precipitation observations by Arizona Weather Stations are accessible at \url{https://www.arcgis.com/home/index.html}.} Compared to the Han River dataset, the US lake dataset is relatively simple, consisting of three sites (lakes) and eleven variables. Figure \ref{fig: us_lakes} presents an area covered by the US lakes dataset.
\begin{table}[t]
    \centering
    \caption{Performance comparison of InstaTran and benchmark methods. The most favorable outcomes are indicated in bold. The second and third ranks are underlined and double-underlined, respectively.}
    \begin{adjustbox}{width=\textwidth}
    \begin{tabular}{c c c c c c c} 
    \Xhline{1.5pt}  
    Metric & $q$ & ETS & ARIMA & Theta & LightGBM & STA-LSTM \\ \hline 
            \multirow{3}{*}{average $q$\mbox{-level QL}} & 0.9 & 1.040 & 0.826 & 0.896 & \dunderline{0.112} & 0.208 \\
    & 0.5 & 0.696 & 0.644 & 0.702 & 0.276 & 0.535 \\
    & 0.1 & 0.180 & 0.179 & 0.189 & 0.135 & 0.279 \\ \hline
    \multirow{3}{*}{$q$-Rate ($|q - q\mbox{-}\mbox{Rate}|$)} & 0.9 & 0.127 (0.773) &  0.237 (0.663) &  0.158 (0.742) &  \dunderline{0.885 (0.043)} & \underline{0.929 (0.029)} \\
    & 0.5 & 0.097 (0.403) &  0.115 (0.385) &  0.093 (0.407) &  \dunderline{0.474 (0.108)} &  0.475 (0.165) \\
    & 0.1 & 0.067 (0.056) & \underline{0.055 (0.049)} & 0.036 (0.064) & \dunderline{0.059 (0.054)} & \textbf{0.131 (0.041)} \\
    \Xhline{1.5pt}   
    \end{tabular}
    \end{adjustbox}
    
    \bigskip
    \begin{adjustbox}{width=\textwidth}

    \begin{tabular}{ c c c c c c c} 
    \Xhline{1.5pt}  
    Metric & $q$ & HSDSTM & DeepAR & MQ-RNN & TFT & InstaTran \\ \hline 
            \multirow{3}{*}{average $q$\mbox{-level QL}} & 0.9 & 1.943 & 0.184 & 0.144 & \underline{0.108} & \textbf{0.106} \\
    & 0.5 & 1.483 & \textbf{0.166} & \dunderline{0.221} & \underline{0.204} & 0.229 \\
    & 0.1 & 1.528 & 0.149 & \textbf{0.097} & \underline{0.099} & \dunderline{0.123} \\ \hline
    \multirow{3}{*}{$q$-Rate ($|q - q\mbox{-}\mbox{Rate}|$)} & 0.9 & 0.897 (0.052) & 0.446 (0.454) & 0.793 (0.175) & 0.911 (0.086) & \textbf{0.887 (0.026)} \\
    & 0.5 & 0.521 (0.225) &  \underline{0.445 (0.095)} & 0.580 (0.146) & 0.458 (0.254) & \textbf{0.550 (0.052)} \\
    & 0.1 & 0.178 (0.098) & 0.445 (0.345) & 0.085 (0.143) & 0.032 (0.068) & 0.150 (0.075) \\
    \Xhline{1.5pt}   
    \end{tabular}
    \end{adjustbox}
    \label{tab: us_lakes_results}
\end{table}

We set the water level of Havasu Lake, the most downstream of the three lakes, as the target variable, with parameters $B=24, \tau=4$ and $\mathcal{Q} =\{0.1, 0.5, 0.9 \}$. Using a 9-year moving window, the period from 2005 to 2013 is split into three disjoint sets: the training set (the first 4 years), the validation set (the middle 2 years), and the testing set (the final 3 years). A 9-year window is then rolled forward by 3 years, refitting and evaluating the models in the same manner on the period from 2008 to 2016. This procedure is repeated four times up to 2022.
For InstaTran, we implemented a causal structure based on the spatial relations of the sites using simple physical models, specifically adhering to the flow from upstream to downstream.

Table \ref{tab: us_lakes_results} presents the experimental results obtained with the US lake dataset. 
Due to the simplicity of the US lake dataset, InstaTran utilized relatively simple causal relations, yet it still yielded competitive results, particularly in the 0.9-level QL and the 0.9 and 0.5-Rate metrics. These results suggest that our proposed model can serve as a strong benchmark, even without prior knowledge.

\end{document}